\documentclass[conference]{IEEEtran}

\IEEEoverridecommandlockouts
\usepackage{amsmath,amssymb,amsfonts}
\usepackage{algorithmic}
\usepackage{graphicx}
\usepackage{textcomp}
\usepackage{xcolor}
\usepackage{comment}
\usepackage{float}
\usepackage{listings}
\usepackage{tikz}
\usetikzlibrary{trees}
\usepackage{dblfloatfix}
\usepackage{booktabs} 
\usepackage{multirow}
\usepackage{hyperref}
\usepackage{tikz}
\usetikzlibrary{positioning}
\usetikzlibrary{fit}
\usetikzlibrary{backgrounds}
\usetikzlibrary{shapes.geometric}
\usetikzlibrary{shapes.misc}
\usetikzlibrary{arrows.meta}
\usetikzlibrary{calc}
\usepackage{pgfplots}

\def\BibTeX{{\rm B\kern-.05em{\sc i\kern-.025em b}\kern-.08em
    T\kern-.1667em\lower.7ex\hbox{E}\kern-.125emX}}

\begin{document}

\title{Intelligent Framework for Human-Robot Collaboration: Dynamic Ergonomics and Adaptive Decision-Making}

\author{\IEEEauthorblockN{Francesco Iodice\IEEEauthorrefmark{1}}
\IEEEauthorblockA{Email: francesco.iodice@polimi.it}
\and
\IEEEauthorblockN{Elena De Momi\IEEEauthorrefmark{1}}
\IEEEauthorblockA{ Email: elena.demomi@polimi.it}
\and
\IEEEauthorblockN{Arash Ajoudani\IEEEauthorrefmark{2}}
\IEEEauthorblockA{Email: arash.ajoudani@iit.it}
\and
\IEEEauthorblockA{ 
\IEEEauthorrefmark{1}{\small Department of Electronics, Information, and Bioengineering, Politecnico di Milano, Milano, Italy.} \and \hspace{1.5cm}
\IEEEauthorrefmark{2}{\small Human-Robot Interfaces and physical Interaction HRI$^{2}$ Lab of Istituto Italiano di Tecnologia (IIT), Genoa, Italy.}
}
}
\maketitle

\begin{table*}[t]
\centering
\renewcommand{\arraystretch}{1.0} 
\setlength{\tabcolsep}{6pt}      
\fontsize{8}{10}\selectfont     
\begin{tabular}{|c|c|c|c|c|c|}
\hline
\textbf{Model} & \textbf{Params (M)} & \textbf{FLOP (G)} & \textbf{mAPval 50-95 (\%)} & \textbf{Latency (ms)} & \textbf{Segmentation} \\
\hline
YOLOv8-N & 3.2 & 8.7 & 37.3 & 6.16 & Yes (YOLO-seg) \\
YOLOv8-S & 11.2 & 28.6 & 44.9 & 7.07 & Yes (YOLO-seg) \\
YOLOv8-M & 25.9 & 78.9 & 50.6 & 9.50 & Yes (YOLO-seg) \\
YOLOv8-L & 43.7 & 165.2 & 52.9 & 12.39 & Yes (YOLO-seg) \\
YOLOv8-X & 68.2 & 257.8 & 53.9 & 16.86 & Yes (YOLO-seg) \\
YOLOv9t & 2.0 & 7.7 & 38.3 & N/A & Yes (YOLO-seg) \\
YOLOv9s & 7.2 & 26.7 & 46.8 & N/A & Yes (YOLO-seg) \\
YOLOv9m & 20.1 & 76.8 & 51.4 & N/A & Yes (YOLO-seg) \\
YOLOv9c & 25.5 & 102.8 & 53.0 & N/A & Yes (YOLO-seg) \\
YOLOv9e & 58.1 & 192.5 & 55.6 & N/A & Yes (YOLO-seg) \\
YOLOv10-N & 2.3 & 6.7 & 39.5 & 1.84 & No \\
YOLOv10-S & 7.2 & 21.6 & 46.8 & 2.49 & No \\
YOLOv10-M & 15.4 & 59.1 & 51.3 & 4.74 & No \\
YOLOv10-L & 24.4 & 120.3 & 53.4 & 7.28 & No \\
YOLOv10-X & 29.5 & 160.4 & 54.4 & 10.70 & No \\
YOLO11-N & 2.6 & 6.5 & 39.5 & 1.50 & Yes (YOLO-seg) \\
YOLO11-S & 9.4 & 21.5 & 47.0 & 2.50 & Yes (YOLO-seg) \\
YOLO11-M & 20.1 & 68.0 & 51.5 & 4.70 & Yes (YOLO-seg) \\
YOLO11-L & 25.3 & 86.9 & 53.4 & 6.20 & Yes (YOLO-seg) \\
YOLO11-X & 56.9 & 194.9 & 54.7 & 11.30 & Yes (YOLO-seg) \\
EfficientDet (D3) & 21.5 & 46.0 & 45.4 & 3.82 & No \\
Faster R-CNN & 41.3 & 87.5 & 49.8 & 6.38 & No \\
Mask R-CNN & 44.5 & 170.0 & 52.0 & 12.8 & Yes \\
\hline
\end{tabular}
\vspace{0.1cm} 
\caption{\textbf{Performance comparison of object detection models on COCO dataset.} The table presents model complexity (Parameters in millions, FLOPs in billions), detection accuracy (mean Average Precision at IoU thresholds 0.5-0.95), processing speed (inference latency in milliseconds), and segmentation capabilities. YOLO models show progressive improvements across generations, with YOLO11-L achieving 53.4\% mAP at 6.20ms latency while maintaining segmentation functionality, demonstrating the trade-offs between computational requirements and performance metrics relevant for real-time applications.}

\label{tab:comparison_coco}
\end{table*}

\begin{abstract}
The integration of collaborative robots into industrial environments has improved productivity, but has also highlighted significant challenges related to operator safety and ergonomics. This paper proposes an innovative framework that integrates advanced visual perception, continuous ergonomic monitoring, and adaptive Behaviour Tree decision-making to overcome the limitations of traditional methods that typically operate as isolated components. Our approach synthesizes deep learning models, advanced tracking algorithms, and dynamic ergonomic assessments into a modular, scalable, and adaptive system. Experimental validation demonstrates the framework's superiority over existing solutions across multiple dimensions: the visual perception module outperformed previous detection models with 72.4\% mAP@50:95; the system achieved high accuracy in recognizing operator intentions (92.5\%); it promptly classified ergonomic risks with minimal latency (0.57 seconds); and it dynamically managed robotic interventions with exceptionally responsive decision-making capabilities (0.07 seconds), representing a 56\% improvement over benchmark systems. This comprehensive solution provides a robust platform for enhancing human-robot collaboration in industrial environments by prioritizing ergonomic safety, operational efficiency, and real-time adaptability.
\end{abstract}

\begin{IEEEkeywords}
Human-robot collaboration, Real-time ergonomics, Visual perception, Adaptive decision-making, Intention recognition, Integrated safety framework
\end{IEEEkeywords}

\section{Introduction}

Industrial automation has revolutionized manufacturing processes and enabled the integration of collaborative robotic systems, allowing robots and human operators to work together and enhancing the productivity and adaptability of production lines. This paradigm, termed Human-Robot Collaboration (HRC), has shown considerable productivity enhancements while also presenting intricate issues concerning safety, ergonomics, and flexibility in dynamic operational situations \cite{pedrocchi2013safe}.
\begin{figure}[t]
    \centering
    \includegraphics[width=\columnwidth]{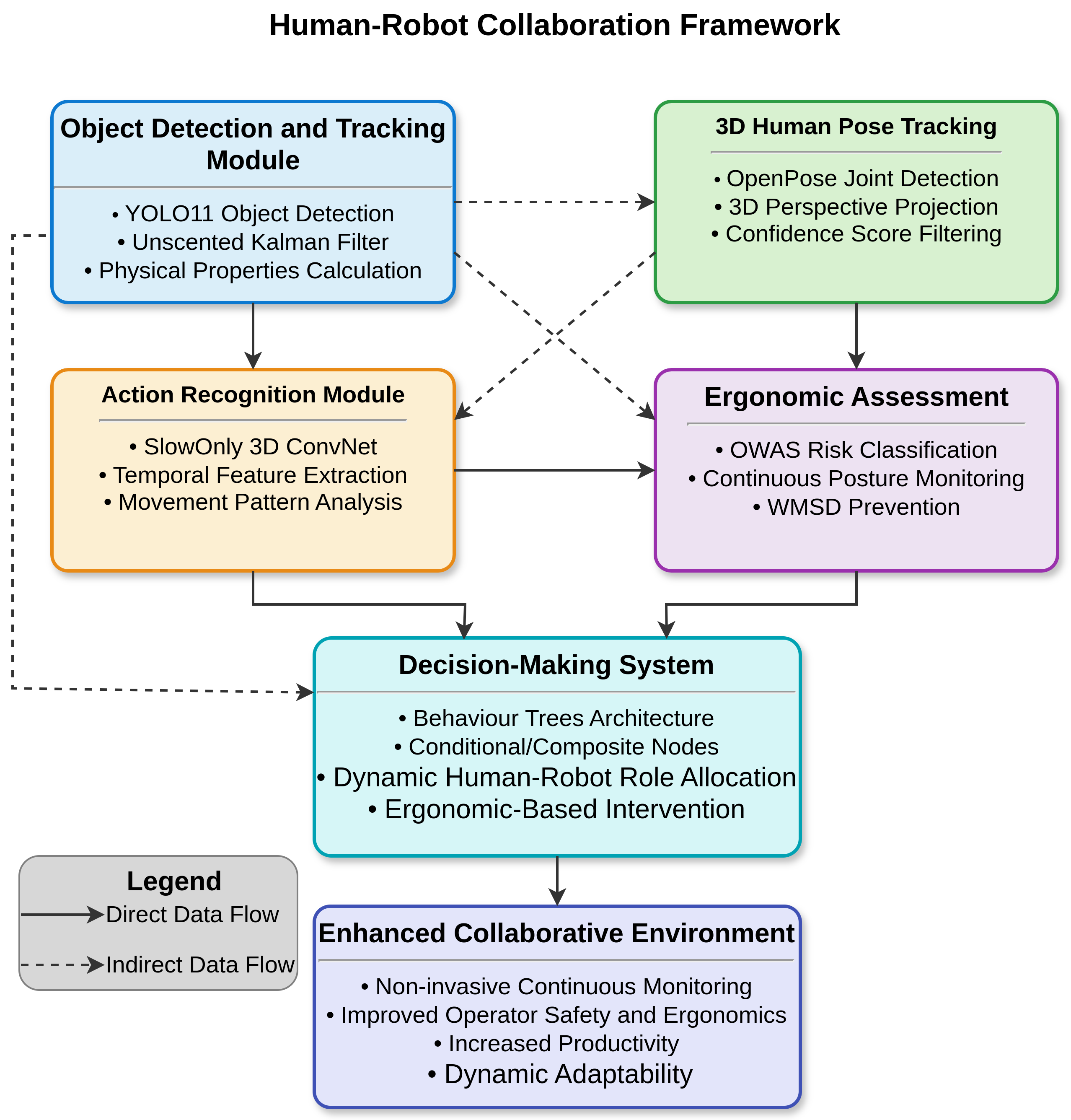}
        \caption{Proposed HRC framework integrating object detection and tracking (YOLO11, Unscented Kalman Filter), 3D human pose tracking (OpenPose), action recognition (SlowOnly), OWAS-based ergonomic assessment, and a Behaviour Trees decision-making system. Solid arrows represent direct data flows between primary functional components, while dashed arrows indicate indirect or complementary information exchanges that support the main processing pipeline. The architecture facilitates non-invasive continuous monitoring, dynamic human-robot role allocation, and ergonomic-based interventions to enhance safety, ergonomics, and productivity in industrial environments.}
    \label{fig:framework}
\end{figure}
The prevention of Work-Related Musculoskeletal Disorders (WMSD) \cite{rahman2023emerging}, which can arise due to incorrect postures or repetitive strain, is one of the most relevant issues. These disorders not only compromise the health of operators, but also negatively affect company productivity, increasing operating costs. As demonstrated by extensive studies on muscle fatigue in work environments \cite{Mahdavi2020fatigue}, these issues require comprehensive monitoring solutions that can adapt to dynamic working conditions and provide timely interventions. Traditional methodologies such as the Ovako Working Posture Analysis System (OWAS) and Rapid Entire Body Assessment (REBA) offer useful tools for identifying ergonomic risks, but are inadequate for the continuous monitoring required in today's complex industrial environments.
To the best of our knowledge, no existing framework in the literature synergistically integrates advanced technologies such as visual detection, continuous ergonomic monitoring, and adaptive decision-making through Behaviour Trees (BT) for collaborative industrial settings. Current approaches address these challenges in isolation or only partially, limiting their practical applicability \cite{villani2018survey}. This work aims to bridge this gap by proposing an innovative framework that combines state-of-the-art technologies to enhance safety, ergonomics, and efficiency in industrial environments.
The proposed framework, illustrated in Fig. \ref{fig:framework}, stands out for its non-invasive nature and its ability to adapt to complex and dynamic operational scenarios. It integrates advanced visual detection technologies (YOLO11 with Unscented Kalman Filter tracking and physical properties calculation, and OpenPose with 3D perspective projection) for comprehensive posture and object recognition, an action recognition module using temporal analysis for movement pattern interpretation, a modular decision-making system based on BTs with conditional and composite nodes for dynamic human-robot role allocation and ergonomic-based intervention, and continuous OWAS-based ergonomic assessment methods to prevent WMSD and other physical risk situations. This integration approach addresses the safety challenges highlighted in industrial collaboration studies \cite{pedrocchi2013safe}, while providing the adaptability required for varied manufacturing tasks and the non-invasiveness that traditional sensor-based systems \cite{conforti2020measuring} often lack. This unified approach enables continuous monitoring and optimization of human-robot interactions, enhancing both safety and overall productivity.

The remainder of this paper is organized as follows: Section II reviews the key related works to highlight existing gaps and motivate our approach. Section III provides a detailed description of the proposed framework, with a focus on its technological modules and system architecture. Section IV presents the experimental results, while Section V discusses potential future developments of the system.

\section{Related Works}
In recent years, the field of human-robot collaboration has seen significant developments in several areas, such as visual perception, ergonomic evaluation and decision-making models to ensure safe and efficient robotic intervention.  As highlighted by Villani et al. \cite{villani2018survey}, these advancements have created new possibilities for intuitive and safe human-robot interactions in industrial settings, yet the integration of these technologies remains a significant challenge. Despite these advances, current solutions often suffer from limited integration of available technologies or take a static approach to ergonomics and safety. The need for more dynamic and adaptative approaches has been emphasized in multiple studies on work-related musculoskeletal disorders \cite{Mahdavi2020fatigue, rahman2023emerging}, which identify real-time monitoring and adaptation as key factors in preventing occupational injuries.

\subsection{Visual Perception and Object Detection}
Visual perception constitutes the foundational layer of human-robot collaborative systems, determining the quality of all subsequent processing stages. The evolution of object detection methodologies reveals a critical transition from accuracy-prioritizing two-stage detectors to efficiency-oriented single-stage architectures designed for time-sensitive applications.
 Early two-stage detection methods, such as R-CNN \cite{girshick2014rich} and Fast R-CNN \cite{ren2015faster}, achieved high accuracy but were computationally inefficient. Faster R-CNN \cite{ren2015faster} introduced region proposal networks to reduce computational requirements; however, their latency remains relatively high, limiting real-time application potential in industrial contexts (see Table \ref{tab:comparison_coco}).

Single-stage detection frameworks, initiated by the YOLO architecture \cite{redmon2016yolo}, significantly reduced latency by formulating detection as a regression problem. Subsequent iterations addressed specific limitations: YOLOv2 \cite{redmon2017yolo9000} improved localization accuracy, YOLOv3 \cite{redmon2018yolov3} enhanced multi-scale detection, and YOLOv4 \cite{bochkovskiy2020yolov4} optimized performance across platforms. More recently, YOLOv8 incorporated segmentation capabilities (YOLO-seg), enabling more detailed scene understanding critical for industrial tasks. YOLOv9 introduced advanced attention mechanisms improving feature representation, whereas YOLOv10 prioritized latency optimization, compromising segmentation capabilities. YOLO11 balanced these trade-offs, offering enhanced segmentation performance and low latency, making it highly suitable for real-time HRC applications, as highlighted by metrics in Table \ref{tab:comparison_coco}.

Alternative architectures like EfficientDet \cite{tan2020efficientdet} explored compound scaling to efficiently balance model complexity and detection accuracy, achieving good speed but lacking native segmentation capabilities. Mask R-CNN \cite{he2017mask} extended Faster R-CNN by incorporating instance segmentation, providing high accuracy but at significant computational costs and latency unsuitable for real-time scenarios (Table \ref{tab:comparison_coco}).

Detection performance degradation under real-world industrial conditions such as occlusions and variable lighting was demonstrated by Dodge and Karam \cite{dodge2016understanding}, and Michaelis et al. \cite{michaelis2019benchmarking}. YOLO11 addresses these limitations effectively, as it maintains high performance in variable industrial environments, which our experimental validation confirmed.

\subsection{Real-Time Tracking}

Real-time tracking is crucial to ensure safety and fluidity in human-robot interactions. The Speed and Separation Monitoring (SSM) method proposed by Marvel et al. \cite{marvel2015speed} monitors the speed and distance between human and robot to prevent collisions, but the lack of integration with advanced visual perception technologies limits its fluidity in complex environments . In contrast, Simple Online and Realtime Tracking (SORT) \cite{wojke2017simple} offers a more responsive solution, using Kalman filtering \cite{kalman1960filter} for rapid detection coupling, but suffers from reduced performance in the presence of occlusions or non-linear movements, which are common challenges in dynamic industrial environments.

More recent developments, such as DeepSORT and ByteTrack \cite{zhang2022bytetrack}, improve robustness under occlusion conditions. Our proposed approach integrates YOLO11 with an Unscented Kalman Filter (UKF) \cite{wan2000unscented}, extending traditional Kalman filter principles to effectively manage complex human movements typical of industrial environments.

\subsection{Pose Detection and Action Recognition}

Human pose and action recognition is essential for improving safety and efficiency in human-robot collaborations. While OpenPose \cite{cao2017realtime} offers powerful pose estimation capabilities, its integration with ergonomic analysis systems remains challenging, as noted in human-robot collaboration surveys \cite{villani2018survey}. Although OpenPose \cite{cao2017realtime} is widely used for real-time human joint detection, it does not offer dynamic ergonomic evaluation or decision-making based on classified actions.

In our previous work \cite{iodice2022hri30}, we demonstrated that the SlowOnly network offers better performance in recognising slow, repetitive movements common in industrial settings than models such as SlowFast \cite{feichtenhofer2019slowfast} and I3D \cite{peng2023i3d}. 

Lasota et al. \cite{lasota2017survey}, in an extensive survey of safety methodologies in human-robot interactions, highlight the importance of human action recognition for collision avoidance. However, most approaches are based on two-dimensional models, which do not adequately address the temporal and spatial complexity of human actions. Our framework, which integrates OpenPose 3D and SlowOnly, offers a more robust and efficient three-dimensional analysis to improve safety and ergonomics.

Cherubini et al. \cite{cherubini2016collaborative} explore collision avoidance in production scenarios, but do not integrate a three-dimensional system for action recognition, nor neural networks optimised for temporal recognition. In contrast, our approach exploits SlowOnly to accurately recognise human actions, improving the management of physical interactions in real time. This attention to temporal patterns in human movement aligns with the safety priorities outlined in collaborative manufacturing studies \cite{pedrocchi2013safe, villani2018survey}, which identify action prediction as a key component of proactive safety systems. Peternel et al. \cite{peternel2018fatigue} propose a system for the management of muscle fatigue during human-robot collaborations, but do not exploit networks optimised for the detection of complex and repetitive movements. Our framework fills this gap, improving ergonomic adaptation and robotic response through more sophisticated action recognition capabilities that can detect subtle movement patterns associated with fatigue and ergonomic risk \cite{Mahdavi2020fatigue}.

\subsection{Dynamic Role Allocation with Behaviour Trees}

Dynamic role allocation is essential for efficient collaboration, especially in industrial settings characterized by rapidly changing conditions. Traditional methods, such as probabilistic models based on human demonstrations proposed by Rozo et al. \cite{rozo2016role}, often exhibit limitations in adapting flexibly to dynamic environments. Similarly, Finite State Machines (FSMs) face scalability challenges due to rapidly increasing complexity \cite{lamon2019capability}, and Markov Decision Processes (MDPs) \cite{iovino2022survey} often require extensive computational resources, limiting their practical application in real-time adaptive scenarios.

To overcome these limitations, our framework employs Behaviour Trees (BT) \cite{colledanchise2018behavior}, which provide a modular, hierarchical, and adaptable decision-making structure. Behaviour Trees facilitate rapid adaptation of robotic tasks by continuously monitoring operational conditions. Compared to probabilistic models, FSMs, and MDPs, BTs significantly simplify the integration of multiple real-time input streams, enhancing responsiveness and adaptability.

Recent advancements in unified architectures for dynamic role allocation \cite{unifiedArchitectureRoleAllocation2023} underline the importance of such adaptable frameworks. Merlo et al. \cite{merlo2023ergonomic} have also proposed ergonomic-driven role allocation aimed at reducing musculoskeletal fatigue. Our work further develops this concept by integrating continuous ergonomic monitoring, significantly improving overall safety, flexibility, and operational responsiveness in dynamically evolving industrial environments \cite{lasota2017survey}.

\subsection{Real-Time Ergonomic Assessment}

Traditional ergonomic assessment techniques, such as OWAS \cite{karhu1977owas}, RULA \cite{mcatamney1993rula}, REBA \cite{hignett2000reba} and the NIOSH Lifting Equation \cite{waters1993niosh}, are based on manual observations and post-hoc analyses, and are unsuitable for continuous and dynamic monitoring of modern industrial environments. These methods have been systematically compared for their effectiveness in identifying potential work-related musculoskeletal disorders \cite{Kee2021owas}. More recent studies, such as that of Ferraguti et al. \cite{ferraguti2020unified} have proposed a solution to automate ergonomic assessment in HRC collaborations, but such approaches do not always succeed in continuously monitoring the physical condition of operators.

Our approach overcomes these limitations by integrating dynamic ergonomic analysis with advanced computer vision technologies such as OpenPose, enabling continuous monitoring of postures in real time \cite{david2005path}. This real-time analysis capability addresses a fundamental gap identified in traditional ergonomic evaluation methods \cite{karhu1977owas, waters1993niosh}, which typically require manual observation and cannot adapt to rapidly changing work conditions. This approach not only prevents injuries related to incorrect postures as cataloged in traditional ergonomic assessment methods \cite{mcatamney1993rula, hignett2000reba}, but also enables an immediate adaptive response, improving safety and reducing operator muscle fatigue through interventions that align with established ergonomic principles \cite{Kee2021owas}.

\begin{figure*}[t]
\centering
\vspace{10pt} 
\includegraphics[width=1.0\textwidth, keepaspectratio]{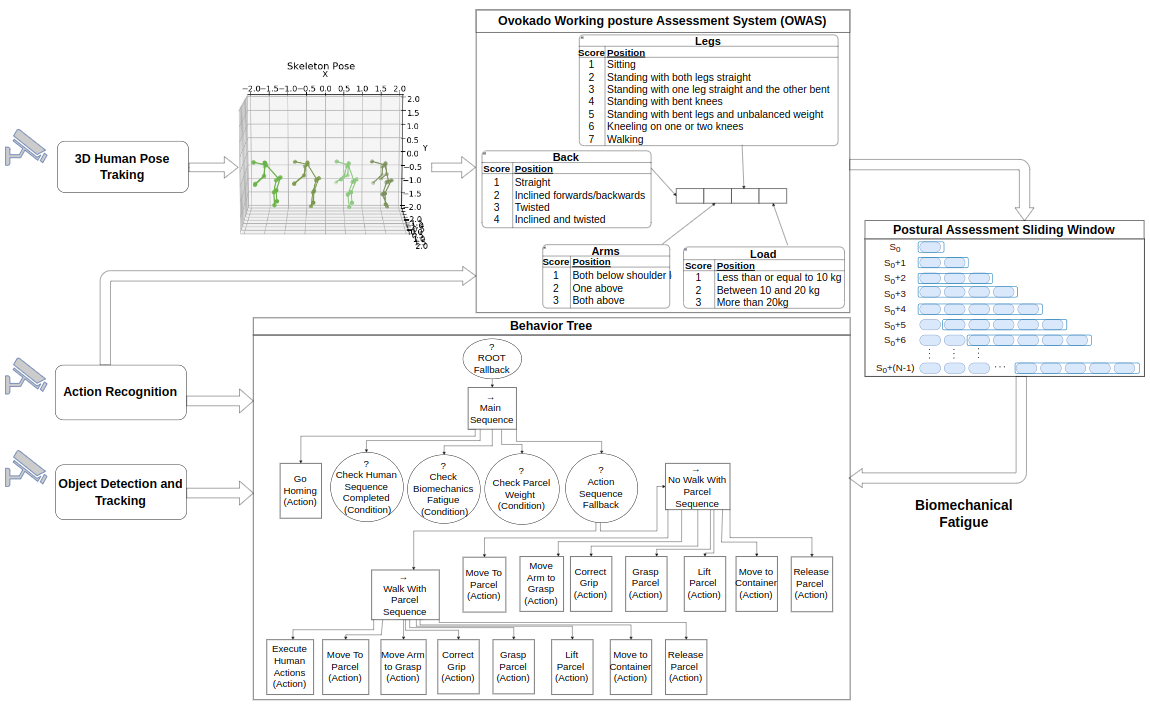} 
\caption{Overview of the proposed human-robot collaboration framework. The architecture integrates three input modules (3D Human Pose Tracking, Action Recognition, and Object Detection and Tracking) that feed into the OWAS ergonomic assessment system. The ergonomic evaluation classifies postures according to back, arms, legs, and load parameters, incorporating a temporal sliding window to track sustained postures. Based on this assessment, the Behaviour Tree manages dynamic role allocation between human operator and robot, prioritizing robot intervention when biomechanical fatigue is detected. This integrated approach ensures operational efficiency while preserving worker ergonomic safety.}
\label{fig:Final_framework}

\end{figure*}

\subsection{Constraints of Wearable Sensor Methodologies and Benefits of Computer Vision}

Several studies have explored the use of wearable sensors to monitor ergonomic risk and assess operators' movements in work contexts. For example, Santopaolo et al. \cite{santopaolo2022biomechanical} used inertial sensors and machine learning to classify biomechanical risks related to lifting, while Donisi et al. \cite{donisi2021work} combined wearable sensors with the NIOSH Lifting Equation to provide a detailed assessment of ergonomic risks in lifting tasks. Conforti et al. \cite{conforti2020measuring} have developed a system based on wearable sensors to monitor operators' movements, demonstrating the effectiveness of these systems for collecting detailed posture and movement data.

Despite their accuracy, wearable sensor-based approaches have significant limitations. Sensors may be invasive, interfering with operators' movements and requiring ongoing management for charging, calibration, and maintenance. Moreover, such systems can increase operational costs, especially in large-scale industrial environments where every worker must be equipped with physical devices. 

In contrast, our framework based on artificial vision offers a non-invasive solution for continuous ergonomic assessment. Using technologies such as YOLO11 \cite{yolov11} for object detection and OpenPose for human posture analysis, the system monitors operators' posture and movements in real-time without requiring the use of physical devices. This approach builds upon established object detection principles \cite{redmon2016yolo} while overcoming the limitations of traditional monitoring systems, providing a scalable solution that can be deployed across multiple workstations without additional hardware costs per operator. This allows a more natural assessment of ergonomic conditions, dynamically adapting to changes in operators' movements.

In addition, computer vision allows for scalable coverage in complex environments, monitoring multiple operators and robots simultaneously without the need for additional sensors. The system can identify incorrect postures or risky movements and intervene in real-time, reducing the risk of repetitive motion or incorrect posture-related injuries, as confirmed by previous studies on musculoskeletal disorders \cite{Mahdavi2020fatigue, waters1993niosh}.

In summary, although wearable sensors offer high accuracy, our computer vision-based approach has significant advantages in terms of flexibility, non-invasiveness and scalability, making it particularly suitable for dynamic industrial settings where real-time ergonomic assessment is required. This approach aligns with the evolution of industrial safety paradigms \cite{villani2018survey, pedrocchi2013safe} toward more integrated and adaptable solutions that can accommodate the full range of human-robot collaborative scenarios while minimizing disruption to existing workflows.

\subsection{Proposed Framework Improvements} 

Compared to previous works, particularly those that address individual aspects of human-robot collaboration \cite{marvel2015speed, cao2017realtime, colledanchise2018behavior} rather than providing an integrated solution, our framework has the following advantages:

\begin{itemize}
\item \textbf{Continuous Ergonomic Monitoring}: Real-time ergonomic assessments improve proactive injury prevention by dynamically identifying high-risk postures and movements.
\item \textbf{Enhanced Action Recognition}: Integration of OpenPose 3D and SlowOnly enables accurate detection of subtle and repetitive actions indicative of fatigue or ergonomic risks.
\item \textbf{Intelligent Task Adaptation}: Utilizing Behaviour Trees facilitates adaptable robot interventions based on real-time conditions, optimizing safety and efficiency.
\end{itemize}

\section{Proposed Framework Architecture}

The proposed framework, illustrated in Fig.~\ref{fig:Final_framework}, combines advanced computer vision techniques, human action recognition, and ergonomic evaluation, enabling synergy between human operator and robot in collaborative environments. This architecture enables continuous monitoring of operators' physical conditions, with an adaptive decision-making system managing dynamic task allocation between human and robot via a BT.

The modular design of our framework allows for component-level improvements and adaptation to various industrial contexts without requiring complete system redesign, addressing a key limitation of monolithic approaches identified in previous research \cite{villani2018survey}. Integration between different technologies, such as computer vision and ergonomic evaluation systems, is essential to improving operations safety and efficiency. In the following, we analyze each key module, explaining how they interact with each other and contribute to the overall operation of the framework.

\subsection{Object Detection and Tracking Module}

This module is responsible for detecting, tracking, and calculating the physical characteristics of objects in the environment, providing essential data for both the decision-making system and ergonomic evaluation.

\subsubsection{Object Detection with YOLO11}

For object detection and the generation of accurate segmentation masks, the YOLO11 model is used, which provides detailed data on the position and size of detected objects, essential for ensuring safety in human-robot collaboration environments. The output of the model is represented by a vector $\mathbf{y}$, which includes position, size, confidence and segmentation of objects:
\[
\mathbf{y} = (x, y, w, h, c, \mathbf{p}, \mathbf{m})
\]

where $(x, y)$ represents the coordinates of the center of the bounding box, $w$ and $h$ are the width and height of the bounding box, $c$ denotes the confidence of the object's presence, $\mathbf{p}$ is the vector of classification probabilities, and $\mathbf{m}$ is the binary segmentation mask.

The model is optimized through a loss function that balances location, confidence, classification, and segmentation:

\[
\mathcal{L}_{\text{totale}} = \mathcal{L}_{\text{box}} + \lambda_{\text{conf}} \mathcal{L}_{\text{conf}} + \lambda_{\text{cls}} \mathcal{L}_{\text{cls}} + \lambda_{\text{seg}} \mathcal{L}_{\text{seg}}
\]

Where:
\begin{itemize}
    \item $\mathcal{L}_{\text{box}}$ represents the bounding box regression loss that penalizes localization errors in the predicted object boundaries
    \item $\mathcal{L}_{\text{conf}}$ represents the confidence loss that quantifies the model's certainty about the presence of an object
    \item $\mathcal{L}_{\text{cls}}$ represents the classification loss that penalizes incorrect category predictions
    \item $\mathcal{L}_{\text{seg}}$ represents the segmentation loss that measures the accuracy of the predicted binary mask
    \item $\lambda_{\text{conf}}$, $\lambda_{\text{cls}}$ and $\lambda_{\text{seg}}$ are scalar weights that balance the relative importance of each component of the loss
\end{itemize}

This multi-objective loss function enables joint optimization of spatial localization and semantic understanding. This formulation allows the system to detect and segment objects in complex scenarios, ensuring accurate separation even in the presence of overlaps or irregular shapes.

\subsubsection{Physical Property Calculation for Ergonomic Assessment}

Using the data provided by YOLO11, the system performs the parallel calculation of the area, volume, and volumetric weight of objects, which are key inputs for ergonomic evaluation using the OWAS methodology.

\paragraph{Area} The area of the object in the 2D projection is calculated as:
\[
A_{\text{object}} = w \cdot h
\]
where $w$ and $h$ are the width and height of the bounding box.

\paragraph{Volume} The volume of the object is estimated by multiplying the area by the average depth $D_{\text{object}}$ detected in the bounding box:
\[
V_{\text{object}} = A_{\text{object}} \cdot (D_{\text{max}} - D_{\text{object}})
\]
where $D_{\text{max}}$ represents the maximum distance and $D_{\text{object}}$ is the average depth detected by the camera RealSense.

\paragraph{Volumetric Weight} Once the volume is estimated, the volumetric weight, $P_{\text{vol}}$, is calculated by dividing the volume by a standard conversion factor:
\[
P_{\text{vol}} = \frac{V_{\text{object}}}{\text{conversion factor}}
\]

where $\text{conversion factor}$ is a standard value of 6000 used in logistics companies to calculate volumetric weight.

This input is critical to the calculation of the OWAS score, where the weight of objects handled by operators is one of the key factors in assessing ergonomic risk.

\subsubsection{Unscented Kalman Filter for Movement Tracking}

In parallel with calculating the physical properties of objects, the UKF is used to track the positions and movements of objects and operators' hands in real time. The UKF is particularly suitable for handling nonlinear dynamics typical of complex movements in industrial environments.

Tracking is handled through a state vector $\mathbf{x} \in \mathbb{R}^{30}$ that includes the positions, velocities, and orientations of objects and hands:

\[
\mathbf{x} = \begin{bmatrix}
\mathbf{p}_{\text{parcel}} \\
\mathbf{v}_{\text{parcel}} \\
\mathbf{q}_{\text{parcel}} \\
\mathbf{p}_{\text{hand\_left}} \\
\mathbf{v}_{\text{hand\_left}} \\
\mathbf{q}_{\text{hand\_left}} \\
\mathbf{p}_{\text{hand\_right}} \\
\mathbf{v}_{\text{hand\_right}} \\
\mathbf{q}_{\text{hand\_right}}
\end{bmatrix}
\]

Here, $\mathbf{p} \in \mathbb{R}^3$ denotes the position in 3D space, $\mathbf{v} \in \mathbb{R}^3$ the linear velocity in 3D space, and $\mathbf{q} \in \mathbb{R}^4$ the orientation expressed in quaternions to avoid the ambiguities that arise in the use of Euler angles.

The state update is conducted through a transition function \(\mathbf{f}: \mathbb{R}^{30} \rightarrow \mathbb{R}^{30}\) that predicts the evolution of the variables over time. This function takes into account the linear and rotational dynamics of the system and is expressed as follows:

\[
\mathbf{f(x)} = \begin{bmatrix}
\mathbf{p}_{\text{parcel}} + \mathbf{v}_{\text{parcel}} \cdot dt \\
\mathbf{v}_{\text{parcel}} \\
\mathbf{q}_{\text{parcel}} + \frac{1}{2} \mathbf{q}_{\text{parcel}} \otimes \mathbf{\Omega}(\boldsymbol{\omega}_{\text{parcel}}) \cdot dt \\
\mathbf{p}_{\text{hand\_left}} + \mathbf{v}_{\text{hand\_left}} \cdot dt \\
\mathbf{v}_{\text{hand\_left}} \\
\mathbf{q}_{\text{hand\_left}} + \frac{1}{2} \mathbf{q}_{\text{hand\_left}} \otimes \mathbf{\Omega}(\boldsymbol{\omega}_{\text{hand\_left}}) \cdot dt \\
\mathbf{p}_{\text{hand\_right}} + \mathbf{v}_{\text{hand\_right}} \cdot dt \\
\mathbf{v}_{\text{hand\_right}} \\
\mathbf{q}_{\text{hand\_right}} + \frac{1}{2} \mathbf{q}_{\text{hand\_right}} \otimes \mathbf{\Omega}(\boldsymbol{\omega}_{\text{hand\_right}}) \cdot dt
\end{bmatrix}
\]

In this formulation:
\begin{itemize}
    \item \( \mathbf{p} \in \mathbb{R}^3 \) is the updated position as a function of linear velocity \( \mathbf{v} \) and time interval \( dt \in \mathbb{R}^+ \),
    \item \( \mathbf{v} \in \mathbb{R}^3 \) represents velocity, held constant within the update interval \( dt \). This choice is justified by the high update rate of the system (30 Hz), which makes speed changes between two consecutive updates negligible,
    \item \( \mathbf{q} \in \mathbb{R}^4 \) represents the updated orientation using the quaternion product \( \otimes \) with the angular velocity \( \boldsymbol{\omega} \in \mathbb{R}^3 \) and time interval \( dt \),
    \item \( \mathbf{\Omega}: \mathbb{R}^3 \rightarrow \mathbb{R}^4 \) is a function that converts angular velocity into quaternion derivative format.
\end{itemize}

The predicted estimate of the state, after updating, is given by the transition function:
\[  
\hat{\mathbf{x}}_{k+1} = \mathbf{f(\mathbf{x}_{k})} 
\]

where $\hat{\mathbf{x}}_{k+1} \in \mathbb{R}^{30}$ represents the predicted state at time step $k+1$, and $\mathbf{x}_{k} \in \mathbb{R}^{30}$ is the current state at time step $k$.

\paragraph{State correction}

After estimating the future state with the transition function, the UKF compares this estimate with actual observations obtained from depth sensors, such as the RealSense D435 camera. 
The measurement function \(\mathbf{h}: \mathbb{R}^{30} \rightarrow \mathbb{R}^{21}\) maps the estimated state onto observables, i.e., the position and orientation of objects and hands:

\[
\mathbf{h(x)} = \begin{bmatrix}
\mathbf{p}_{\text{parcel}} \\
\mathbf{q}_{\text{parcel}} \\
\mathbf{p}_{\text{hand\_left}} \\
\mathbf{q}_{\text{hand\_left}} \\
\mathbf{p}_{\text{hand\_right}} \\
\mathbf{q}_{\text{hand\_right}}
\end{bmatrix}
\]

The measurement vector $\mathbf{z}_k \in \mathbb{R}^{21}$ consists of the direct observations of positions and orientations, excluding velocities which are not directly measurable. The measurement predicted at time $k$, $\hat{\mathbf{z}}_k \in \mathbb{R}^{21}$ is obtained by applying the measurement function to the state predicted at time $k$:

\[
\hat{\mathbf{z}}_k = \mathbf{h}(\hat{\mathbf{x}}_{k})  
\]

Subsequently, the new actual observations \({\mathbf{z}}_k\) are compared with those predicted estimates. The error between the predicted estimate \(\hat{\mathbf{z}}_k\) and the actual observations \(\mathbf{z}_k\) is used to update and correct the estimated state, via the Kalman gain \( K \in \mathbb{R}^{30 \times 21} \). The final correction of the state \(\mathbf{x}_{k+1}\) is made with the following equation:

\[
\mathbf{x}_{k+1} = \hat{\mathbf{x}}_{k+1} + K (\mathbf{z}_k - \hat{\mathbf{z}}_k)
\]

The Kalman gain $K$ is computed internally by the UKF algorithm and optimally weights the relative importance of the model prediction versus the new measurements, based on their respective uncertainties.

\paragraph{Covariance Matrices}

The stability and accuracy of the system strongly depend on a proper definition of the covariance matrices. The process noise covariance matrix \(\mathbf{Q} \in \mathbb{R}^{30 \times 30}\) represents the uncertainty in the system model, while the observation noise matrix \(\mathbf{R} \in \mathbb{R}^{21 \times 21}\) handles the uncertainty associated with sensor measurements:

\[
\mathbf{Q} = \text{diag}([100, \dots, 100])
\quad \text{and} \quad
\mathbf{R} = \text{diag}([0.01, \dots, 0.01])
\]

The initial state covariance matrix \(\mathbf{P} \in \mathbb{R}^{30 \times 30}\), defined as a scaled identity matrix, also plays a key role in the accuracy of the first estimates:

\[
\mathbf{P} = 0.01 \cdot \mathbf{I}_{30}
\]

where $\mathbf{I}_{30} \in \mathbb{R}^{30 \times 30}$ is the identity matrix of dimension 30.

The UKF implementation uses the Merwe scaled sigma point parameterization with $\alpha = 10^{-3}$, $\beta = 2.0$, and $\kappa = 0$, which determines how the sigma points are distributed around the mean to capture the system's nonlinearities. To ensure numerical stability, a small regularization term ($1e^{-8}$) is added to the covariance matrices, and positive definiteness is enforced through eigenvalue decomposition when necessary.

The correct selection of these parameters is essential to ensure that the UKF system maintains stability and accuracy even in the presence of uncertainties in the process or measurements, minimizing estimation error throughout the entire operational cycle.

\begin{figure*}[t]
\centering
\vspace{10pt} 
\includegraphics[width=1.0\textwidth, keepaspectratio]{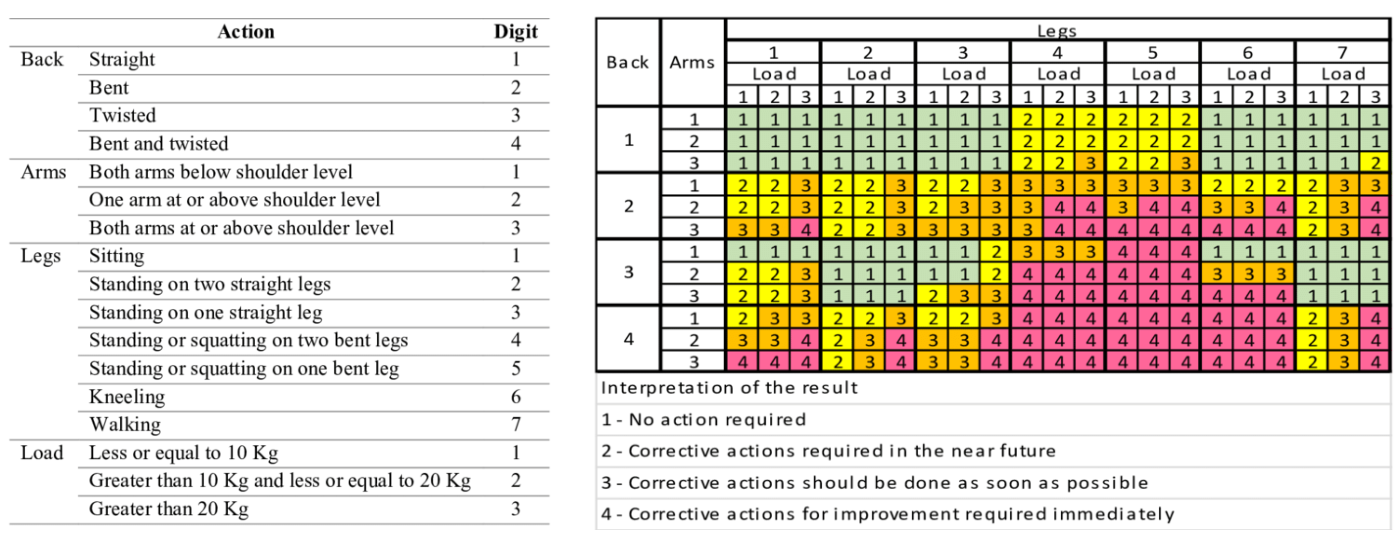} 
\caption{OWAS (Ovako Working Posture Analysis System) evaluation matrix. The left section displays the coding system for back (1-4), arms (1-3), legs (1-7), and load (1-3) postures. The right section presents the action category matrix with color-coded risk levels (green: no action required; yellow: corrective actions required in the near future; orange: corrective actions should be done as soon as possible; red: corrective actions for improvement required immediately) based on combinations of back, arms, legs, and load values.}
\label{fig:OWAS_fig}

\end{figure*}

\subsection{3D Human Pose Tracking Module}
The 3D Human Pose Tracking Module is responsible for detecting and analyzing the operator's posture in real-time, providing crucial data for ergonomic evaluation.

\subsubsection{Pose Detection with OpenPose}

OpenPose detects the joints of the human body in 3D using a perspective projection \cite{iodice2022learning}. Each i-th joint is represented by the coordinates $(x, y, z)$ and a confidence score $c$:

\[
\mathbf{J} = \left\{ (x_i, y_i, z_i, c_i) \right\}, \, i = 1, 2, \dots, N
\]

The collected data are subsequently analyzed by the SlowOnly module.

\subsection{Action Recognition Module}
The Action Recognition Module analyzes the operator's movements over time to identify specific actions and patterns, contributing to both ergonomic assessment and decision-making.

\subsubsection{Temporal Analysis with SlowOnly}

The SlowOnly model, a specialized three-dimensional convolutional network (3D ConvNet), analyzes the slow, repetitive movements of operators. The input is a sequence of video frames $F_{\text{in}}$, sampled at regular temporal intervals:

\[
F_{\text{in}} = \left[ I_t, I_{t+\Delta t}, I_{t+2\Delta t}, \dots, I_{t+T\Delta t} \right]
\]

Where:
\begin{itemize}
    \item $I_t \in \mathbb{R}^{H \times W \times C}$ represents the video frame at time $t$, with $H$ being the height, $W$ the width, and $C$ the number of channels of the frame
    \item $\Delta t \in \mathbb{R}^{+}$ is the temporal sampling interval between consecutive frames
    \item $T \in \mathbb{N}$ is the total number of frames in the sequence
\end{itemize}

Three-dimensional convolution applied to this sequence extracts both spatial and temporal features:

\[
F_{\text{out}} = \text{Conv3D}(F_{\text{in}}, K, S)
\]

Where:
\begin{itemize}
    \item $F_{\text{out}} \in \mathbb{R}^{H' \times W' \times D \times N}$ is the output feature tensor
    \item $K \in \mathbb{R}^{k_h \times k_w \times k_t \times C \times N}$ represents the 3D convolutional kernels
    \item $S \in \mathbb{N}^3$ denotes the stride parameters $(s_h, s_w, s_t)$
\end{itemize}

The temporal features extracted by SlowOnly are then forwarded to the Ergonomic Assessment Module for evaluation of the operator's movements using the OWAS methodology.

\subsection{Ergonomic Assessment Module}
The Ergonomic Assessment Module integrates data from the previous modules to evaluate the operator's posture and working conditions according to established ergonomic standards.

\subsubsection{Ovako Working Posture Analysis System}
The Ovako Working Posture Analysis System (OWAS) classifies operators' postures into risk categories, allowing the system to monitor and intervene when postures are detected that could lead to musculoskeletal injuries.

Fig.~\ref{fig:OWAS_fig} illustrates the OWAS evaluation matrix used in our framework. This standardized tool provides a systematic approach to assessing ergonomic risks through a color-coded matrix that indicates different levels of necessary intervention based on various posture combinations.
\paragraph{Classification of Postures}
The classification of operator postures, as shown in Fig.~\ref{fig:OWAS_fig}, is based on a systematic coding system that evaluates four key body segments:
\begin{enumerate}
    \item \textbf{Posture of the back (1-4)}: 
       \begin{itemize}
         \item (1) Straight
         \item (2) Bent
         \item (3) Twisted
         \item (4) Bent and twisted
       \end{itemize}
    \item \textbf{Posture of the arms (1-3)}: 
       \begin{itemize}
         \item (1) Both arms below shoulder level
         \item (2) One arm at or above shoulder level
         \item (3) Both arms at or above shoulder level
       \end{itemize}
    \item \textbf{Position of legs (1-7)}: 
       \begin{itemize}
         \item (1) Sitting
         \item (2) Standing on two straight legs
         \item (3) Standing on one straight leg
         \item (4) Standing or squatting on two bent legs
         \item (5) Standing or squatting on one bent leg
         \item (6) Kneeling
         \item (7) Walking
       \end{itemize}
    \item \textbf{Weight lifted (1-3)}: 
       \begin{itemize}
         \item (1) Less than or equal to 10 kg
         \item (2) Greater than 10 kg and less than or equal to 20 kg
         \item (3) Greater than 20 kg
       \end{itemize}
\end{enumerate}

The OWAS evaluation process combines these four codes to determine a risk level using the standardized evaluation matrix shown in Fig.~\ref{fig:OWAS_fig}. This matrix cross-references specific combinations of postures and load to assign one of four action categories:
\begin{itemize}
    \item \textbf{Action Category 1}: No action required - normal posture without harmful effect on the musculoskeletal system.
    \item \textbf{Action Category 2}: Corrective actions required in the near future - posture has some harmful effect on the musculoskeletal system.
    \item \textbf{Action Category 3}: Corrective actions should be done as soon as possible - posture has a distinctly harmful effect on the musculoskeletal system.
    \item \textbf{Action Category 4}: Corrective actions for improvement required immediately - posture has an extremely harmful effect on the musculoskeletal system.
\end{itemize}

\begin{figure*}[t]
\centering
\vspace{10pt} 
\includegraphics[width=1.0\textwidth, keepaspectratio]{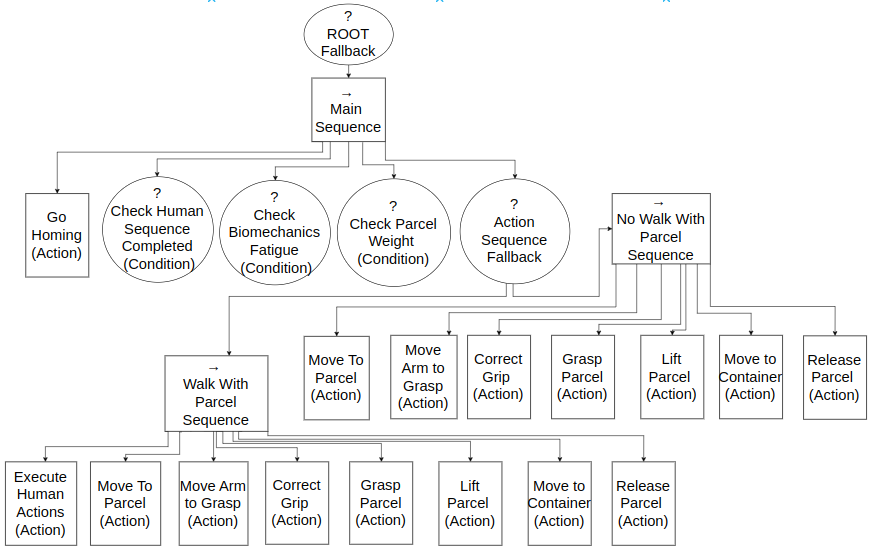} 
\caption{Hierarchical structure of the implemented Behaviour Tree for dynamic role allocation. The tree starts with a ROOT Fallback node and branches through a Main Sequence into conditional checks for robot position, human sequence completion, biomechanical fatigue, and parcel weight. Based on these evaluations, the Action Sequence Fallback node directs execution either to robot-led actions (via No Walk With Parcel Sequence) or human-led operations with robot support (via Walk With Parcel Sequence).}
\label{fig:bt_diagram}

\end{figure*}
\paragraph{OWAS Score Calculation}
While the standard OWAS methodology determines the action category directly through the evaluation matrix, our implementation calculates a continuous weighted score to enable more nuanced ergonomic assessment. The formula for calculating this weighted OWAS score is:

\[
P_{\text{owas}} = \alpha \cdot P_{\text{back}} + \beta \cdot P_{\text{arm}} + \gamma \cdot P_{\text{leg}} + \delta \cdot P_{\text{weight}}
\]

where:
\begin{itemize}
    \item $P_{\text{back}}$, $P_{\text{arm}}$, $P_{\text{leg}}$, $P_{\text{weight}}$ are the numerical codes assigned to each postural category (ranging from 1-4, 1-3, 1-7, and 1-3 respectively)
    \item $\alpha$, $\beta$, $\gamma$, $\delta$ are weighting coefficients that modulate the relative contribution of each postural factor
\end{itemize}

The weighting coefficients ($\alpha$, $\beta$, $\gamma$, $\delta$) are derived from ergonomic research on work-related musculoskeletal disorders. In the original OWAS methodology \cite{karhu1977owas}, these weights reflect the relative impact of different body segments on overall postural strain. 
The exact values are calibrated based on:

\begin{itemize}
    \item The prevalence of disorders in different body regions in industrial settings
    \item The biomechanical load associated with each posture type
    \item The time sustainability of various postures before discomfort onset
    \item The recovery time needed after maintaining specific postures
\end{itemize}

The calculated weighted score is then mapped to the four action categories using threshold values that align with the OWAS evaluation matrix shown in Fig.~\ref{fig:OWAS_fig}. This approach allows our system to continuously monitor ergonomic risk levels and trigger appropriate interventions when hazardous postures are detected.

\paragraph{Framework Integration of Ergonomic Assessment}
The OWAS methodology is integrated directly into the framework to monitor operators' posture in real time, establishing a computational pathway between sensor data and ergonomic risk assessment. The integration process consists of the following steps:
\begin{enumerate}
    \item \textbf{Skeletal data acquisition}: OpenPose detects the 3D coordinates and confidence scores of key body joints, while SlowOnly analyzes temporal movement patterns to identify sustained postures.
    \item \textbf{Posture mapping}: The acquired skeletal data is mapped to the corresponding OWAS coding system (back, arms, legs), while load parameters are derived from the volumetric calculations performed by the Visual Perception Module.
    \item \textbf{OWAS score computation}: The weighted score is calculated using the formula previously described, with the calibrated coefficients reflecting the biomechanical impact of each postural component.
    \item \textbf{Dynamic intervention}: When the computed ergonomic risk exceeds the predefined thresholds corresponding to OWAS action categories, the system initiates appropriate intervention strategies to mitigate the identified hazards.
\end{enumerate}

This approach transforms the traditional manual OWAS assessment into an automated, continuous monitoring system capable of detecting ergonomic hazards in real-time industrial scenarios.

\subsection{Decision-Making with Behaviour Trees}
Within the proposed framework, dynamic role allocation between human operator and robot is managed through a BT. This decision structure allows the system to adapt robotic behaviour based on operational conditions, operator actions, and real-time sensory inputs. Unlike traditional approaches, such as finite-state diagrams or Markovian decision models, the BT offers greater flexibility and modularity, which is particularly relevant in dynamic industrial settings where human-robot collaboration requires fast and accurate decisions.

\subsubsection{Structure and Execution of the Behaviour Tree}
The BT implemented in the framework is hierarchically structured and composed of different types of nodes:
\begin{itemize}
\item \textbf{Action Nodes}: They perform concrete operations, such as "grabbing an object" or "moving to a destination." These nodes allow the robot to perform the required actions based on the detected conditions. 
\item \textbf{Conditional Nodes}: They evaluate whether or not the action can proceed by monitoring parameters such as operator biomechanical fatigue or completion of a sequence of human movements. For example, a conditional node might check whether the lifted package has an acceptable weight for handling. 
\item \textbf{Composite Nodes}: These nodes combine multiple actions and conditions, creating complex logic flows. Composite nodes used include: 
\begin{itemize} 
\item \textit{Sequence}: Nodes are executed in sequential order until all actions are completed correctly. If a node fails, execution stops. 
\item \textit{Fallback}: Nodes are executed sequentially until one succeeds. This node type is particularly useful for handling alternative situations where the primary action is unavailable or fails. 
\end{itemize} 
\end{itemize}

As illustrated in Fig.~\ref{fig:bt_diagram}, the BT structure begins with a ROOT Fallback node at the top level, which provides the initial decision-making capability. This root node connects to a Main Sequence node that orchestrates the operational flow. From the Main Sequence, the tree branches into five parallel nodes that constitute the core decision-making components:
\begin{itemize}
    \item \textit{Go Homing} (Action): Verifies that the robot is in the correct starting position before initiating operations.
    \item \textit{Check Human Sequence Completed} (Condition): Evaluates whether the operator has completed their intended sequence of movements.
    \item \textit{Check Biomechanics Fatigue} (Condition): Monitors operator biomechanical fatigue based on ergonomic data collected in real time.
    \item \textit{Check Parcel Weight} (Condition): Evaluates whether the weight of the package falls within acceptable parameters for handling.
    \item \textit{Action Sequence Fallback}: A composite fallback node that determines the appropriate action path based on the outcome of preceding conditions.
\end{itemize}

The execution process begins at the ROOT Fallback node and proceeds through the Main Sequence. Based on the evaluation of the conditional nodes, the tree implements two primary execution pathways:
\begin{itemize}
    \item The first pathway through the \textit{No Walk With Parcel Sequence} directs robot-led actions when the operator's ergonomic condition indicates risk. This sequence leads to actions including \textit{Move To Parcel}, \textit{Move Arm to Grasp}, \textit{Correct Grip}, \textit{Grasp Parcel}, \textit{Lift Parcel}, \textit{Move to Container}, and \textit{Release Parcel}.
    
    \item The second pathway through the \textit{Walk With Parcel Sequence} manages human-led operations with robot support when ergonomic conditions are favorable. This includes monitoring through \textit{Execute Human Actions} and providing assistance as needed.
\end{itemize}

When the OWAS assessment indicates elevated ergonomic risk (for example, due to improper posture or fatigue detected via the Check Biomechanics Fatigue node), the BT prioritizes robot intervention. Conversely, if the ergonomic assessment indicates favorable conditions for human operation, the tree allows for human-led execution with robot monitoring. This dynamic allocation ensures that workload is distributed optimally based on real-time ergonomic conditions.

\subsubsection{Advantages of the Behaviour Tree Approach}
The adoption of the BT in the implemented framework offers several advantages over other control techniques, including: 
\begin{itemize} 
\item \textbf{Modularity}: Each node in the BT is independent and modifiable, allowing behaviours to be added or removed without affecting the entire system. This enables the framework to be easily extended with new actions or conditions as requirements evolve.

\item \textbf{Flexibility}: The system can dynamically adapt to changes in the operator's physical condition or variations in operational activities. For example, if the human operator shows signs of fatigue or adopts risky postures (as detected by the OWAS assessment), the BT automatically triggers robot intervention to lift and move packages, maintaining operator safety.

\item \textbf{Real-Time Responsiveness}: Through integration with visual perception modules (e.g., YOLO11 and OpenPose) and ergonomic evaluation, the BT enables the robot to make quick and accurate decisions, optimizing both operator safety and operational efficiency. The millisecond-level response capability is essential in dynamic industrial environments where conditions can change rapidly.
\end{itemize}

\begin{table*}[t]
\centering
\caption{Comparative analysis of object detection performance across YOLO model generations}
\label{tab:global_performance}
\begin{tabular}{l|cc}
\hline
\textbf{Model Architecture} & \textbf{mAP@50 (\%)} & \textbf{mAP@50:95 (\%)} \\
\hline
YOLOv8x-seg & 77.6 & 71.2 \\
YOLOv9-seg & 77.7 & 72.4 \\
YOLO11x-seg & \textbf{77.8} & \textbf{72.4} \\
\hline
\multicolumn{3}{p{0.8\textwidth}}{\footnotesize Note: Mean Average Precision metrics at IoU threshold of 0.5 (mAP@50) and across multiple IoU thresholds from 0.5 to 0.95 (mAP@50:95) demonstrate the incremental improvement trajectory across model generations. Best results are highlighted in bold.} \\
\end{tabular}
\end{table*}

\section{Experiments and Results}
To evaluate the overall effectiveness of the proposed framework in human-robot collaboration, experiments were designed and conducted in a controlled laboratory environment and in simulation. The laboratory experiments replicated realistic operational scenarios, focusing on ergonomic monitoring, action detection, visual perception, and overall integration of the framework components.
In parallel, the Behaviour Tree (BT)-based decision-making module was tested in simulation to analyze its dynamic and adaptive management capability in a wide range of complex scenarios that are difficult to replicate in the laboratory.

\subsection{Experimental Setup}

All experiments were conducted using a standardized hardware and software infrastructure to ensure consistency across different test scenarios:

\subsubsection{Hardware Components}
\begin{enumerate}
    \item \textbf{Visual Sensing}: Two Intel RealSense D435 depth cameras were deployed for object detection/tracking and human pose estimation respectively. An additional Logitech C270 webcam was used for capturing video streams for action recognition analysis.
    
    \item \textbf{Robotic System}: The experiments utilized a collaborative robotic arm (Universal Robots UR16) mounted on the Summit XL mobile platform, configured to respond to commands from the decision-making system.
    
    \item \textbf{Computing Resources}: The framework operated on a distributed network comprising four workstations in a ROS-based master-slave configuration. Three workstations were dedicated to vision processing, equipped with different GPUs (NVIDIA GTX 970, NVIDIA GTX 1080Ti, and NVIDIA RTX A3000) to handle various perception tasks. The fourth workstation was the onboard computer of the mobile platform, which managed robot control operations.
\end{enumerate}

\subsubsection{Software Framework}
The system ran on Ubuntu 20.04 with ROS Noetic providing the communication infrastructure between components. The YOLO11 detection module, OpenPose tracking, and SlowOnly action recognition algorithms were deployed across the vision processing nodes according to their computational requirements. Neural network models were initially trained offline, while experiment execution utilized the distributed GPU setup for real-time inference.

This heterogeneous computing architecture was designed to maintain system responsiveness while processing multiple data streams, enabling the ergonomic assessment and decision-making modules to operate effectively in dynamic conditions.

\begin{table*}[t]
\centering
\caption{Class-specific performance evaluation of state-of-the-art YOLO segmentation models}
\label{tab:class_performance}
\begin{tabular}{l|l|ccc}
\hline
\textbf{Class} & \textbf{Metrics} & \textbf{YOLOv8x-seg} & \textbf{YOLOv9-seg} & \textbf{YOLO11x-seg} \\
\hline
\multirow{4}{*}{\textit{Parcel}} 
& Precision (\%) & 88.0 & 85.6 & \textbf{89.8} \\
& Recall (\%) & 79.3 & 80.9 & \textbf{81.1} \\
& mAP@50 (\%) & \textbf{89.4} & \textbf{89.4} & 89.3 \\
& mAP@50:95 (\%) & 86.7 & 86.0 & \textbf{86.8} \\
\hline
\multirow{4}{*}{\textit{Hand}}
& Precision (\%) & \textbf{89.8} & 89.6 & \textbf{89.8} \\
& Recall (\%) & 59.9 & 60.9 & \textbf{61.1} \\
& mAP@50 (\%) & 65.7 & 66.1 & \textbf{66.3} \\
& mAP@50:95 (\%) & 57.3 & 53.2 & \textbf{58.0} \\
\hline
\multicolumn{5}{p{0.8\textwidth}}{\footnotesize Note: Performance metrics illustrate the differential detection capabilities across object classes, with superior performance observed in parcel detection compared to hand detection. YOLO11x-seg demonstrates improvements in critical metrics, particularly in hand detection precision and recall. Best results for each metric are highlighted in bold.} \\
\end{tabular}
\end{table*}

\subsection{Experiment: Grasping Action Detection and Volumetric Weight Estimation with Variable Size Parcels}

\begin{figure*}[t] 
    \centering
    \begin{minipage}[t]{0.48\textwidth}
        \centering
        \includegraphics[width=\textwidth]{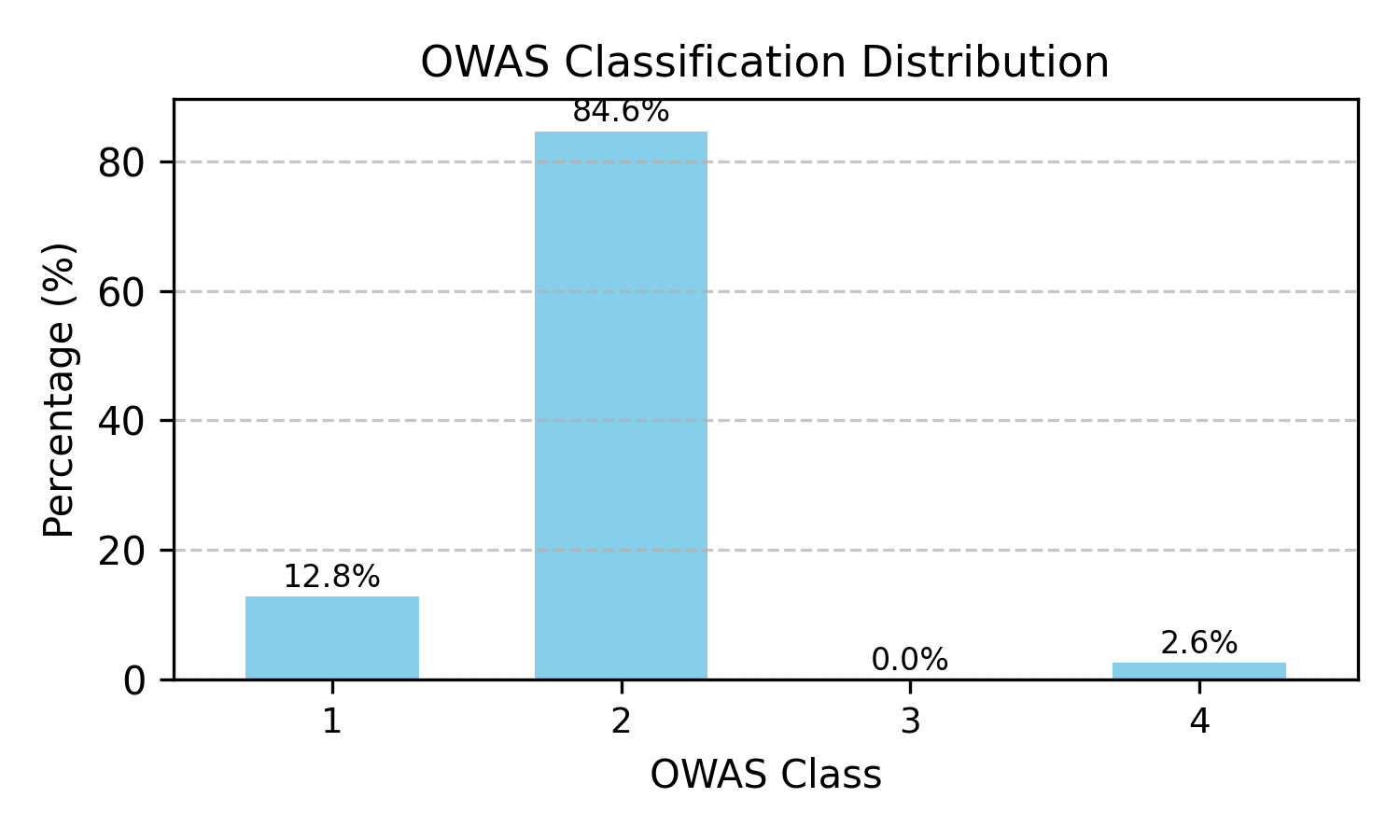}
        \caption{Distribution of postures classified according to OWAS classes during the experiment. Most of the postures belong to OWAS class 2 (moderate risk), with a small percentage in OWAS 4 (high risk).}
        \label{fig:OWAS_classification_Distribution}
    \end{minipage}%
    \hfill
    \begin{minipage}[t]{0.48\textwidth}
        \centering
        \includegraphics[width=\textwidth]{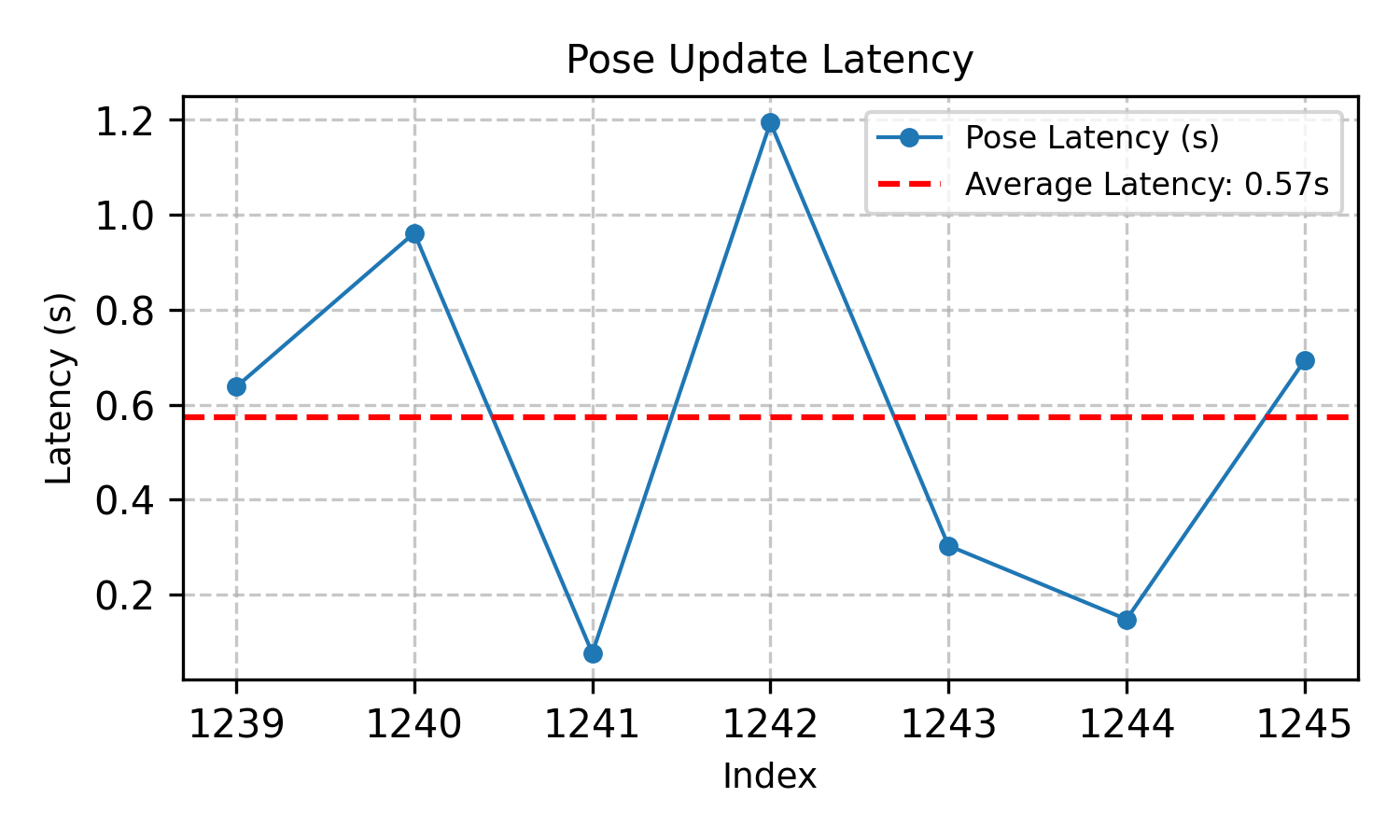}
        \caption{Update latencies of monitored poses. An average latency of 0.57 seconds ensures timely detections and supports rapid interventions.}
        \label{fig:Pose_Update_Latencies}
    \end{minipage}
\end{figure*}

In this preliminary phase of the framework, the main objective was the recognition of grasping intention and estimation of volumetric weight of parcels of varying sizes. The dataset used included 15,160 images annotated with the Segment Anything Model (SAM), ensuring accurate segmentation of hands and parcels into diverse configurations.Images were divided into training (80\%), validation (10\%) and testing (10\%), ensuring a balanced distribution for robust evaluation.

Three segmentation models were analyzed: YOLOv8x-seg, YOLOv9-seg and YOLO11x-seg, trained with the same configuration parameters and evaluated on two main classes, \textit{"parcel"} and \textit{"hand"}. 

For grasping intention recognition, 40 trials were conducted on dynamic sequences, combining the segmentation model with the probabilistic Kalman Unscented (UKF) filter. For volumetric estimation, 4 parcels of varying sizes and shapes were tested, each in 10 trials for a total of 40 measurements. The estimated volumetric weights, obtained from the segmented masks and depth data provided by the RealSense D435 sensor, were compared with the actual weights.

The performance of the segmentation models was evaluated using mAP@50 (mean Average Precision with an IoU threshold of 0.5) and mAP@50:95 (mean Average Precision averaged over multiple IoU thresholds from 0.5 to 0.95 with a step size of 0.05). Precision, defined as the ratio of true positives to the sum of true positives and false positives, and recall, defined as the ratio of true positives to the sum of true positives and false negatives, were calculated for class-specific analysis alongside mAP values.

For volumetric weight estimation, percentage errors were calculated using the formula:
\[
E_p = \frac{|P_{v,\text{st}} - P_{v,\text{re}}|}{P_{v,\text{re}}} \times 100
\]

where \( P_{v,\text{st}} \) e \( P_{v,\text{re}} \) represent estimated and actual volumetric weights, respectively.

As shown in Table~\ref{tab:global_performance}, YOLOv8x-seg achieved a mAP@50 of 77.6\% and a mAP@50:95 of 71.2\%, demonstrating a good balance between precision and recall. YOLOv9-seg slightly improved the mAP@50 (77.7\%) and maintained the mAP@50:95 at 72.4\%, with more efficient latency management. However, YOLO11x-seg outperformed both, achieving a mAP@50 of 77.8\% and a mAP@50:95 of 72.4\%, showing greater robustness in complex scenarios.

Table~\ref{tab:class_performance} analyzes performance by class. For the class \textit{"parcel"}, YOLOv9-seg and YOLO11x-seg produced similar results, with YOLO11x-seg slightly outperforming YOLOv9-seg in terms of recall and mAP@50:95 (86.8\% versus 86.0\%).  For the class \textit{"hand"}, YOLO11x-seg stood out significantly, achieving a mAP@50:95 of 58.0\%,compared with 53.2\%for YOLOv9-seg and 57.3\% for YOLOv8x-seg. This result highlights the ability of YOLO11x-seg to handle pose variability, occlusions, and structural complexity.

For grasping intention recognition, the system correctly identified grasping in 37 cases out of 40, with an overall accuracy of 92.5\%. The overall mean percent error for volumetric weight estimation over all trials was calculated as 17.58\%, with greater variability observed in the extreme size parcels, attributable to the sensor's limitations in terms of spatial resolution and surface reflectivity.

In conclusion, the model comparison confirmed that YOLO11x-seg represents the best performing solution for the operational context considered. Its combination of robustness, accuracy, and ability to handle difficult conditions, such as detection of moving or partially occluded hands, makes it ideal for grasping and volumetric estimation applications. The experimental outcomes reveal the advantages of integrating a deep learning model with probabilistic tracking algorithms, demonstrating superior robustness compared to the isolated use of neural networks. However, volumetric weight estimation showed sensitivity to the quality of depth data provided by the sensor, with a reduction in accuracy due to limitations in spatial resolution or surface reflectivity. This suggests that further improvements could be achieved through refinement of sensor calibration or integration of data from multiple sensors to compensate for current limitations.
\begin{figure*}[t] 
    \centering
    \begin{minipage}[t]{0.48\textwidth}
        \centering
        \includegraphics[width=\textwidth]{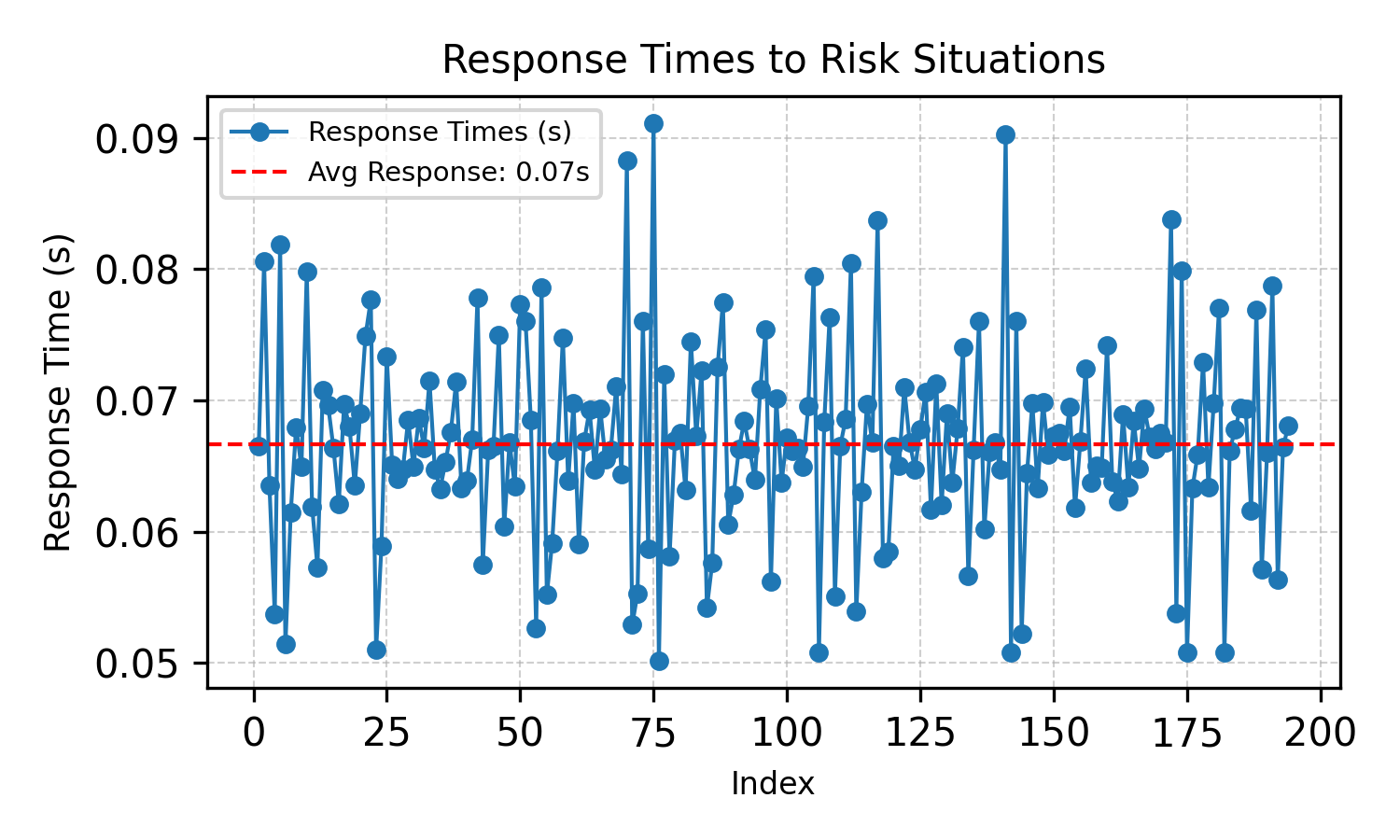}
        \caption{Average response times of the framework to critical conditions. The fast response (0.07 seconds on average) highlights the system's efficiency in making real-time decisions.}
        \label{fig:Response_Times}
    \end{minipage}%
    \hfill
    \begin{minipage}[t]{0.48\textwidth}
        \centering
        \includegraphics[width=\textwidth]{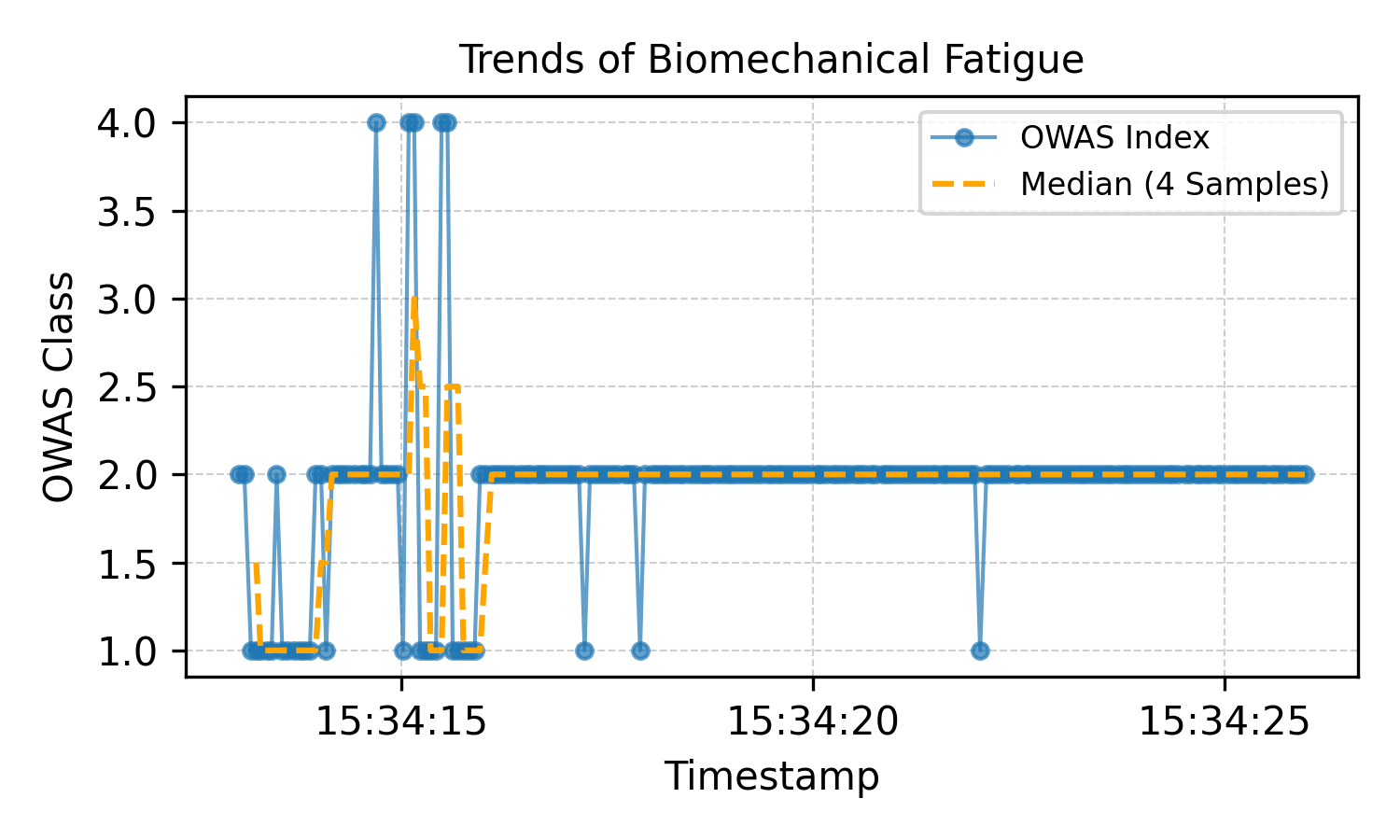}
        \caption{Time trend of the median of the analyzed poses according to OWAS. Predominance of moderate risk (OWAS class 2) with peaks in OWAS class 4 during heavy lifting.}
        \label{fig:OWAS_Trends_Median}
    \end{minipage}
\end{figure*}
\subsection{Experiment: Ergonomic Monitoring, Risk Classification and Robotic Intervention via Behaviour Tree}

The objective of this experiment is to evaluate the framework's ability to monitor operator postures in real time, classify ergonomic risk, and manage task transfer to the robot in the presence of critical conditions. During testing, the operator performed movements typical of an industrial environment, such as bending and lifting, while the system monitored posture using OpenPose and classified ergonomic risk using the OWAS method.

When medium-high-risk (OWAS class 3) or high-risk (OWAS class 4) postures were detected, or volumetric weight in excess of allowable limits, the system generated a biomechanical fatigue message. This message was processed by a BT, which combined ergonomic data, action recognition results, and volumetric weight values to determine whether robotic intervention should be activated. This approach ensured modular and responsive decision making, allowing a smooth transition between the operator and the robot.

The performance of this experiment was evaluated through several metrics: the distribution of postures across OWAS risk classes (1-4), system update rate (Hz) and latency (seconds) for pose monitoring, response time (seconds) to risky situations, and time trend analysis of posture risk levels. 

The data collected during the experiment showed that 12,8\%  of the postures belonged to OWAS class 1 (no risk), 84,6\%  belonged to OWAS class 2 (moderate risk), no postures were classified in OWAS class 3 (medium-high risk), and 2,6\% belonged to OWAS class 4 (high risk), as shown in Fig.~\ref{fig:OWAS_classification_Distribution}. These results confirm the system's ability to distinguish between ergonomically safe and hazardous postures, demonstrating the effectiveness of the framework in monitoring and classifying risk. Compared to the approach based on OpenPose by Lin et al. \cite {lin2022automatic}, our system demonstrates significantly higher precision in ergonomic evaluation. The classification of high-risk postures (OWAS class 4) stands at 2.6\%, drastically reducing false positives compared to the 10.4\% reported in the comparative work. This enhanced diagnostic precision minimizes unnecessary operational interruptions, a crucial aspect in dynamic industrial environments.

Temporally, the system maintained an average update rate of 14.99 Hz, with an average latency of 0.57 seconds for pose monitoring, as shown in Fig.~\ref{fig:Pose_Update_Latencies}. These values ensured that any postural changes were detected and analyzed in a timely manner. In addition, the average response time to risky situations was 0.07 seconds, as shown in Fig.~\ref{fig:Response_Times}, demonstrating the efficiency of the framework in identifying critical conditions and responding quickly. The system's reactivity, measured through the average response time to ergonomic risk situations, reaches 0.07 seconds, exceeding by more than 56\% the benchmark of 0.16 seconds reported by Tortora et al. \cite{tortora2019fast}. This substantial improvement in latency was achieved while maintaining a recognition accuracy of 92.5\%, comparable to the 94.3\% of the reference system.
The optimal balance between speed and precision enables our framework to implement real-time intervention capabilities previously unattainable in dynamic industrial environments, where even a few milliseconds can determine the effectiveness of safety measures.

An analysis of the postures over time windows of four samples, Fig.~\ref{fig:OWAS_Trends_Median} revealed that the median of the postures remained in OWAS class 2 (moderate risk), with temporary peaks in OWAS class 4 during particularly heavy lifting or postures held for a long time. This underscores the system's ability to accurately detect critical moments, providing useful decision support for operations management.

These measurements demonstrate not only the framework's efficiency in detecting and analyzing critical conditions, but also its role in supporting a seamless operational transition. The reduced overall latency, as calculated through pose monitoring and response time to risky situations, ensured that the transfer of activities to the robot was smooth and responsive. This level of operational readiness, combined with the system's ability to integrate ergonomic and actionable data, represents a significant step toward optimizing human-robot collaboration.

\subsection{Experiment: Recognizing Actions with SlowOnly Model}

This experiment evaluated the framework's ability to recognize and classify actions performed by an operator in a simulated industrial environment by exploiting the SlowOnly model. The model, based on a ResNet3D network with SlowOnly architecture and 50-level depth, was pre-trained on the HRI30 dataset, specifically developed to recognize actions in human-robot interaction scenarios. During the experiment, the model was configured to classify three classes of actions: lifting, carrying, and repositioning packages.

Video sequences were preprocessed to 8-frame clips, uniformly sampled with a time interval of four frames, and normalized using ImageNet mean values and standard deviations. Each clip was subjected to resizing, center cropping, and augmentation operations with random color changes and horizontal flips to improve the robustness of the model.
The optimization process used the AdamW algorithm with an initial learning rate of 0.001, adjusted through a Cosine Annealing strategy with warmup for the first 1,000 steps. Training was conducted on an NVIDIA RTX 3090 GPU, using a batch size of 16 videos per GPU for 200 epochs, with the cross-entropy loss function.

The experiment's performance was assessed through classification loss during training, top-1 accuracy on the validation set, comparison with alternative architectures (I3D and TSM), and model robustness to occlusions and variability.

During the validation phase, the model achieved an accuracy of 95,83\%
demonstrating an excellent ability to generalize over the test data. The behaviour of loss during training shows rapid initial convergence, with stabilization at 0.129 toward the later epochs
. This result reflects the effectiveness of the model setup and training pipeline in gradually reducing the classification error. 

The use of short 8-frame clips proved particularly effective in capturing action movements without introducing significant computational complexity, while integration with the BT ensured that the classified data were exploited for real-time robotic decisions.

This model and framework configuration was further validated by the previous evaluation on HRI30, where SlowOnly was shown to outperform alternative architectures such as I3D and TSM in action recognition in industrial settings. Comparative analysis \cite{iodice2022hri30} revealed that SlowOnly achieved a top-1 accuracy of 86.55\% compared to 78.43\% for I3D and 73.91\% for TSM when evaluated on the HRI30 dataset.  The model's robustness to occlusions and its high accuracy make it an ideal choice for collaborative human-robot scenarios, where accurate and timely action recognition is crucial.

Experimental outcomes establish the effectiveness of the proposed approach, which combines action recognition, ergonomic analysis, and volumetric evaluation, integrating them into a modular and responsive decision-making framework. The framework showed a remarkable ability to adapt to operational variables, providing decisive support for safety and efficiency in collaborative industrial operations.\\

\section{Conclusions and Future Work}.

In this work, an innovative framework for human-robot collaboration was proposed, combining advanced visual sensing technologies, real-time ergonomic monitoring, and Behaviour Tree (BT)-based adaptive decision-making. Experimental results show that the system significantly improves operator safety and efficiency of operations, providing a scalable and modular solution for complex collaborative environments.

To the best of our knowledge, this is the first framework to synergistically integrate advanced visual perception, real-time ergonomic assessment, and adaptive decision-making for industrial HRC. Compared with traditional methods based on static rules or post-hoc ergonomic analysis, our approach introduces greater flexibility and adaptability, improving both the safety and well-being of operators. The noninvasive nature of the system, which eliminates the need for wearable physical sensors, also helps to reduce operational costs and improve the naturalness of human-robot interaction.

Despite the encouraging results, the work has some limitations. The experiments, conducted in a controlled laboratory and in simulation, need to be extended to real industrial scenarios to evaluate the robustness of the system under varying operating conditions. In particular, the Behaviour Tree module requires practical validation to demonstrate its effectiveness in dynamic human-robot role management. In addition, the absence of direct comparisons with established technologies presents an opportunity to further position the framework in the state of the art.

To address the limitations that have emerged and broaden the impact of the framework, future work will focus on:

\begin{itemize} \item Validation in Real Environments: Conducting field tests in operational industrial settings to demonstrate the robustness and scalability of the system on a large scale. \item Comparisons with Existing Solutions: Conduct quantitative comparisons with established academic technologies using standard benchmarks to further validate the framework's contribution. \item End-User Engagement: Collaborate with industrial operators and supervisors to gather feedback on the usability, efficiency, and acceptability of the framework, improving its configuration based on specific user needs. \item Expansion for Complex Scenarios: Extend the system to handle multi-operator and multi-robot interactions, improving collaborative efficiency in dynamic and high-variability environments. \item Integration of Advanced Sensors: Experiment with next-generation sensors, such as high-resolution RGB-D cameras and multi-sensor configurations, to further improve the accuracy of visual monitoring and volumetric estimation. 

\item Cross-domain Applications: Explore the framework's adaptability in non-industrial contexts such as healthcare (surgical assistance and patient mobility support), smart homes (assistive living for elderly individuals), rehabilitation centers (personalized therapy tracking), and logistics (warehouse automation and order fulfillment).
\end{itemize}

In conclusion, the proposed framework represents a major breakthrough in human-robot collaboration, combining safety, efficiency and adaptability in a single integrated solution. Its modular and scalable architecture makes it particularly suitable for a wide range of industrial applications, including logistics, manufacturing, and healthcare. With validation in real-world environments and enrichment with additional functionality, the system has the potential to become a benchmark solution, improving the productivity and well-being of operators in modern manufacturing settings.

\bibliography{biblio}

\begin{thebibliography}{10}
\providecommand{\url}[1]{#1}
\csname url@samestyle\endcsname
\providecommand{\newblock}{\relax}
\providecommand{\bibinfo}[2]{#2}
\providecommand{\BIBentrySTDinterwordspacing}{\spaceskip=0pt\relax}
\providecommand{\BIBentryALTinterwordstretchfactor}{4}
\providecommand{\BIBentryALTinterwordspacing}{\spaceskip=\fontdimen2\font plus
\BIBentryALTinterwordstretchfactor\fontdimen3\font minus
  \fontdimen4\font\relax}
\providecommand{\BIBforeignlanguage}[2]{{%
\expandafter\ifx\csname l@#1\endcsname\relax
\typeout{** WARNING: IEEEtran.bst: No hyphenation pattern has been}%
\typeout{** loaded for the language `#1'. Using the pattern for}%
\typeout{** the default language instead.}%
\else
\language=\csname l@#1\endcsname
\fi
#2}}
\providecommand{\BIBdecl}{\relax}
\BIBdecl

\bibitem{pedrocchi2013safe}
N.~Pedrocchi, F.~Vicentini, A.~Tosatti, and L.~Molinari-Tosatti, ``Safe
  human-robot cooperation in an industrial environment,'' \emph{IEEE/ASME
  Transactions on Mechatronics}, vol.~19, no.~1, pp. 151--160, 2013.

\bibitem{rahman2023emerging}
M.~H. Rahman, A.~Ghasemi, F.~Dai, and J.~Ryu, ``Review of emerging technologies
  for reducing ergonomic hazards in construction workplaces,''
  \emph{Buildings}, vol.~13, no.~12, p. 2967, 2023.

\bibitem{Mahdavi2020fatigue}
N.~Mahdavi \emph{et~al.}, ``A review of work environment risk factors
  influencing muscle fatigue,'' \emph{International Journal of Industrial
  Ergonomics}, vol.~80, p. 103028, 2020.

\bibitem{villani2018survey}
V.~Villani, F.~Pini, F.~Leali, and C.~Secchi, ``Survey on human-robot
  collaboration in industrial settings: Safety, intuitive interfaces and
  applications,'' \emph{IEEE Robotics and Automation Letters}, vol.~3, no.~1,
  pp. 1201--1208, 2018.

\bibitem{conforti2020measuring}
I.~Conforti \emph{et~al.}, ``Measuring biomechanical risk in lifting load tasks
  through wearable system and machine-learning approach,'' \emph{Sensors},
  vol.~20, no.~6, p. 1557, 2020.

\bibitem{girshick2014rich}
R.~Girshick, J.~Donahue, T.~Darrell, and J.~Malik, ``Rich feature hierarchies
  for accurate object detection and semantic segmentation,'' in
  \emph{Proceedings of the IEEE conference on computer vision and pattern
  recognition}, 2014, pp. 580--587.

\bibitem{ren2015faster}
S.~Ren, K.~He, R.~Girshick, and J.~Sun, ``Faster r-cnn: Towards real-time
  object detection with region proposal networks,'' in \emph{Advances in Neural
  Information Processing Systems}, vol.~28, 2015, pp. 91--99.

\bibitem{redmon2016yolo}
J.~Redmon, S.~Divvala, R.~Girshick, and A.~Farhadi, ``You only look once:
  Unified, real-time object detection,'' in \emph{IEEE Conference on Computer
  Vision and Pattern Recognition (CVPR)}, 2016, pp. 779--788.

\bibitem{redmon2017yolo9000}
J.~Redmon and A.~Farhadi, ``Yolo9000: better, faster, stronger,'' in
  \emph{Proceedings of the IEEE conference on computer vision and pattern
  recognition}, 2017, pp. 7263--7271.

\bibitem{redmon2018yolov3}
------, ``Yolov3: An incremental improvement,'' \emph{arXiv preprint
  arXiv:1804.02767}, 2018.

\bibitem{bochkovskiy2020yolov4}
A.~Bochkovskiy, C.-Y. Wang, and H.-Y.~M. Liao, ``Yolov4: Optimal speed and
  accuracy of object detection,'' \emph{arXiv preprint arXiv:2004.10934}, 2020.

\bibitem{tan2020efficientdet}
M.~Tan, R.~Pang, and Q.~V. Le, ``Efficientdet: Scalable and efficient object
  detection,'' \emph{arXiv preprint arXiv:2004.01655}, 2020.

\bibitem{he2017mask}
K.~He, G.~Gkioxari, P.~Dollár, and R.~Girshick, ``Mask r-cnn,'' in \emph{IEEE
  International Conference on Computer Vision}, 2017, pp. 2961--2969.

\bibitem{dodge2016understanding}
S.~Dodge and L.~Karam, ``Understanding how image quality affects deep neural
  networks,'' in \emph{2016 eighth international conference on quality of
  multimedia experience (QoMEX)}.\hskip 1em plus 0.5em minus 0.4em\relax IEEE,
  2016, pp. 1--6.

\bibitem{michaelis2019benchmarking}
C.~Michaelis, B.~Mitzkus, R.~Geirhos, E.~Rusak, O.~Bringmann, A.~S. Ecker,
  M.~Bethge, and W.~Brendel, ``Benchmarking robustness in object detection:
  Autonomous driving when winter is coming,'' \emph{arXiv preprint
  arXiv:1907.07484}, 2019.

\bibitem{marvel2015speed}
J.~Marvel, R.~Norcross, and J.~Falco, ``Implementing speed and separation
  monitoring in collaborative robot workcells,'' \emph{IEEE Transactions on
  Automation Science and Engineering}, vol.~12, no.~3, pp. 969--976, 2015.

\bibitem{wojke2017simple}
N.~Wojke, A.~Bewley, and D.~Paulus, ``Simple online and realtime tracking with
  a deep association metric,'' \emph{arXiv preprint arXiv:1703.07402}, 2017.

\bibitem{kalman1960filter}
R.~E. Kalman, ``A new approach to linear filtering and prediction problems,''
  \emph{Journal of Basic Engineering}, vol.~82, no.~1, pp. 35--45, 1960.

\bibitem{zhang2022bytetrack}
Y.~Zhang, P.~Sun, Y.~Jiang, D.~Yu, F.~Weng, Z.~Yuan, P.~Luo, W.~Liu, and
  X.~Wang, ``Bytetrack: Multi-object tracking by associating every detection
  box,'' in \emph{European conference on computer vision}.\hskip 1em plus 0.5em
  minus 0.4em\relax Springer, 2022, pp. 1--21.

\bibitem{wan2000unscented}
E.~A. Wan and R.~Van Der~Merwe, ``The unscented kalman filter for nonlinear
  estimation,'' in \emph{Proceedings of the IEEE Adaptive Systems for Signal
  Processing, Communications, and Control Symposium}, 2000, pp. 153--158.

\bibitem{cao2017realtime}
Z.~Cao, T.~Simon, S.-E. Wei, and Y.~Sheikh, ``Realtime multi-person 2d pose
  estimation using part affinity fields,'' in \emph{IEEE Conference on Computer
  Vision and Pattern Recognition (CVPR)}, 2017, pp. 1302--1310.

\bibitem{iodice2022hri30}
F.~Iodice, E.~De~Momi, and A.~Ajoudani, ``Hri30: An action recognition dataset
  for industrial human-robot interaction,'' in \emph{26th International
  Conference on Pattern Recognition (ICPR)}, 2022, pp. 4941--4947.

\bibitem{feichtenhofer2019slowfast}
C.~Feichtenhofer, H.~Fan, J.~Malik, and K.~He, ``Slowfast networks for video
  recognition,'' in \emph{IEEE International Conference on Computer Vision
  (ICCV)}, 2019, pp. 2004--2013.

\bibitem{peng2023i3d}
Y.~Peng, J.~Lee, and S.~Watanabe, ``I3d: Transformer architectures with
  input-dependent dynamic depth for speech recognition,'' in \emph{IEEE
  International Conference on Acoustics, Speech and Signal Processing
  (ICASSP)}, 2023, pp. 1--5.

\bibitem{lasota2017survey}
P.~A. Lasota, T.~Fong, and J.~A. Shah, ``A survey of methods for safe
  human-robot interaction,'' \emph{Annual Review of Control, Robotics, and
  Autonomous Systems}, vol.~1, pp. 123--149, 2017.

\bibitem{cherubini2016collaborative}
A.~Cherubini, R.~Passama, A.~Crosnier, A.~Lasnier, and P.~Fraisse,
  ``Collaborative manufacturing with physical human-robot interaction,''
  \emph{IEEE Transactions on Automation Science and Engineering}, vol.~13,
  no.~1, pp. 118--129, 2016.

\bibitem{peternel2018fatigue}
L.~Peternel, N.~G. Tsagarakis, and D.~G. Caldwell, ``Robot adaptation to human
  physical fatigue in human-robot co-manipulation,'' \emph{Autonomous Robots},
  vol.~42, no.~5, pp. 1011--1021, 2018.

\bibitem{rozo2016role}
L.~Rozo, P.~Jimenez, C.~Torras, and G.~Alenya, ``Learning physical
  collaborative robot behaviors from human demonstrations,'' \emph{IEEE
  Transactions on Robotics}, vol.~32, no.~3, pp. 513--527, 2016.

\bibitem{lamon2019capability}
E.~Lamon, A.~De~Franco, L.~Peternel, and A.~Ajoudani, ``A capability-aware role
  allocation approach to industrial assembly tasks,'' \emph{IEEE Robotics and
  Automation Letters}, vol.~4, no.~4, pp. 3378--3385, 2019.

\bibitem{iovino2022survey}
M.~Iovino, E.~Scukins, J.~Styrud, P.~{\"O}gren, and C.~Smith, ``A survey of
  behavior trees in robotics and ai,'' \emph{Robotics and Autonomous Systems},
  vol. 154, p. 104096, 2022.

\bibitem{colledanchise2018behavior}
M.~Colledanchise and P.~Ögren, \emph{Behavior Trees in Robotics and AI: An
  Introduction}.\hskip 1em plus 0.5em minus 0.4em\relax CRC Press, 2018.

\bibitem{unifiedArchitectureRoleAllocation2023}
E.~Lamon, F.~Fusaro, E.~De~Momi, and A.~Ajoudani, ``A unified architecture for
  dynamic role allocation and collaborative task planning in mixed human-robot
  teams,'' \emph{arXiv preprint arXiv:2301.08038}, 2023.

\bibitem{merlo2023ergonomic}
E.~Merlo, E.~Lamon, F.~Fusaro, M.~Lorenzini, A.~Carfì, F.~Mastrogiovanni, and
  A.~Ajoudani, ``An ergonomic role allocation framework for dynamic human-robot
  collaborative tasks,'' \emph{Journal of Manufacturing Systems}, vol.~67, pp.
  111--121, 2023.

\bibitem{karhu1977owas}
O.~Karhu, P.~Kansi, and I.~Kuorinka, ``Correcting working postures in industry:
  A practical method for analysis,'' \emph{Applied Ergonomics}, vol.~8, no.~4,
  pp. 199--201, 1977.

\bibitem{mcatamney1993rula}
L.~McAtamney and E.~N. Corlett, ``Rula: A survey method for the investigation
  of work-related upper limb disorders,'' \emph{Applied Ergonomics}, vol.~24,
  no.~2, pp. 91--99, 1993.

\bibitem{hignett2000reba}
S.~Hignett and L.~McAtamney, ``Rapid entire body assessment (reba),''
  \emph{Applied Ergonomics}, vol.~31, no.~2, pp. 201--205, 2000.

\bibitem{waters1993niosh}
T.~R. Waters, V.~Putz-Anderson, A.~Garg, and L.~J. Fine, ``Revised niosh
  equation for the design and evaluation of manual lifting tasks,''
  \emph{Ergonomics}, vol.~36, no.~7, pp. 749--776, 1993.

\bibitem{Kee2021owas}
D.~Kee, ``Comparison of owas, rula and reba for assessing potential
  work-related musculoskeletal disorders,'' \emph{International Journal of
  Industrial Ergonomics}, vol.~83, p. 103140, 2021.

\bibitem{ferraguti2020unified}
F.~Ferraguti, R.~Villa, C.~T. Landi, A.~M. Zanchettin, P.~Rocco, and C.~Secchi,
  ``A unified architecture for physical and ergonomic human-robot
  collaboration,'' \emph{Robotica}, vol.~38, no.~4, pp. 669--683, 2020.

\bibitem{david2005path}
G.~David, ``Ergonomic methods for assessing exposure to risk factors for
  work-related musculoskeletal disorders,'' \emph{Applied Ergonomics}, vol.~36,
  no.~4, pp. 463--473, 2005.

\bibitem{santopaolo2022biomechanical}
A.~Santopaolo, M.~Lorenzini, L.~Privitera, T.~Varrecchia, G.~Chini,
  A.~Ranavolo, P.~Ariano, and A.~Ajoudani, ``Biomechanical risk assessment of
  human lifting tasks via supervised classification of multiple sensor data,''
  in \emph{2022 IEEE-RAS 21st International Conference on Humanoid Robots
  (Humanoids)}.\hskip 1em plus 0.5em minus 0.4em\relax IEEE, 2022, pp.
  746--751.

\bibitem{donisi2021work}
L.~Donisi \emph{et~al.}, ``Work-related risk assessment according to the
  revised niosh lifting equation: A preliminary study using a wearable inertial
  sensor and machine learning,'' \emph{Sensors}, vol.~21, no.~8, p. 2593, 2021.

\bibitem{yolov11}
R.~Khanam and M.~Hussain, ``Yolov11: An overview of key architectural
  enhancements,'' \emph{arXiv preprint arXiv:2410.17725}, 2024.

\bibitem{iodice2022learning}
F.~Iodice, Y.~Wu, W.~Kim, F.~Zhao, E.~De~Momi, and A.~Ajoudani, ``Learning
  cooperative dynamic manipulation skills from human demonstration videos,''
  \emph{Mechatronics}, vol.~85, p. 102807, 2022.

\bibitem{lin2022automatic}
P.-C. Lin, Y.-J. Chen, W.-S. Chen, and Y.-J. Lee, ``Automatic real-time
  occupational posture evaluation and select corresponding ergonomic
  assessments,'' \emph{Scientific Reports}, vol.~12, no.~1, p. 2139, 2022.

\bibitem{tortora2019fast}
S.~Tortora, S.~Michieletto, F.~Stival, and E.~Menegatti, ``Fast human motion
  prediction for human-robot collaboration with wearable interface,'' in
  \emph{2019 IEEE International Conference on Cybernetics and Intelligent
  Systems (CIS) and IEEE Conference on Robotics, Automation and Mechatronics
  (RAM)}.\hskip 1em plus 0.5em minus 0.4em\relax IEEE, 2019, pp. 457--462.

\end{thebibliography}

\end{document}